\documentclass{article}

\usepackage{microtype}
\usepackage{graphicx}
\usepackage{subfigure}
\usepackage{booktabs} %

\usepackage{hyperref}

\usepackage[accepted]{icml2025}

\usepackage{amsmath}
\usepackage{amssymb}
\usepackage{mathtools}
\usepackage{amsthm}

\usepackage[capitalize,noabbrev]{cleveref}
\usepackage{graphicx}
\usepackage{makecell}

\usepackage{comment}
\usepackage{tikz}
\usetikzlibrary{calc} 
\newcommand{\ctikzbar}[3]{
 \begin{tikzpicture}[fill fraction/.style n args={2}{path picture={
 \fill[#1] (path picture bounding box.south west) rectangle
 ($(path picture bounding box.north west)!#2!(path picture bounding box.north
 east)$);}}]

 \draw (0, 0) node[draw, fill fraction={#2}{#1}] {#3};
 \end{tikzpicture}
}
\newcommand{\onone}{\ctikzbar{black!50}{0}{~~~}}
\newcommand{\omin}{\ctikzbar{black!50}{.1}{~~~}}
\newcommand{\olow}{\ctikzbar{black!50}{0.3}{~~~}}
\newcommand{\omed}{\ctikzbar{black!50}{0.6}{~~~}}
\newcommand{\ohigh}{\ctikzbar{black!50}{1.0}{~~~}}

\usepackage{hyperref}
\usepackage{pifont}%
\newcommand{\cmark}{\ding{51}}%
\newcommand{\xmark}{\ding{55}}
\usepackage{soul}
\usepackage{booktabs}  
\usepackage{graphicx}  

\usepackage{soul}

\definecolor{ForestGreen}{RGB}{34,139,34}

\newcommand{\circledtext}[1]{%
    \,\,%
    \tikz[baseline=(char.base)]{
        \node[shape=circle, draw=black, inner sep=1pt] (char) {#1};
    }\,%
}
\theoremstyle{plain}

\theoremstyle{definition}

\theoremstyle{remark}

\usepackage[textsize=tiny]{todonotes}
\usepackage{multirow}

\begin{document}

\twocolumn[
\icmltitle{Position: Editing Large Language Models Poses Serious Safety Risks}

\icmlsetsymbol{equal}{*}

\begin{icmlauthorlist}
\icmlauthor{Paul Youssef}{marburg}
\icmlauthor{Zhixue Zhao}{sheffield}
\icmlauthor{Daniel Braun}{marburg}
\icmlauthor{Jörg Schlötterer}{marburg,mannheim}
\icmlauthor{Christin Seifert}{marburg}
\end{icmlauthorlist}

\icmlaffiliation{marburg}{Marburg University, Marburg, Germany}
\icmlaffiliation{sheffield}{University of Sheffield, Sheffield, UK}
\icmlaffiliation{mannheim}{University of Mannheim, Mannheim, Germany}

\icmlcorrespondingauthor{Zhixue Zhao}{zhixue.zhao@sheffield.ac.uk}

\icmlkeywords{Machine Learning, ICML}

\vskip 0.3in
]

\printAffiliationsAndNotice{}  %
\begin{abstract}

Large Language Models (LLMs) contain large amounts of facts about the world. These facts can become outdated over time, which has led to the development of knowledge editing methods (KEs) that can change specific facts in LLMs with limited side effects. This position paper argues that editing LLMs poses serious safety risks that have been largely overlooked. First, we note the fact that KEs are widely available, computationally inexpensive, highly performant, and stealthy makes them an attractive tool for malicious actors. Second, we discuss malicious use cases of KEs, showing how KEs can be easily adapted for a variety of malicious purposes. Third, we highlight vulnerabilities in the AI ecosystem that allow unrestricted uploading and downloading of updated models without verification. Fourth, we argue that a lack of social and institutional awareness exacerbates this risk, and discuss the implications for different stakeholders. We call on the community to (i) research tamper-resistant models and countermeasures against malicious model editing, and (ii) actively engage in securing the AI ecosystem.

\end{abstract}

\section{Introduction}

\begin{figure}[h]
\centering
     \begin{picture}(300,130)
         \includegraphics[width=\columnwidth, trim={0cm 17.5cm 0cm 0.5cm}, clip]{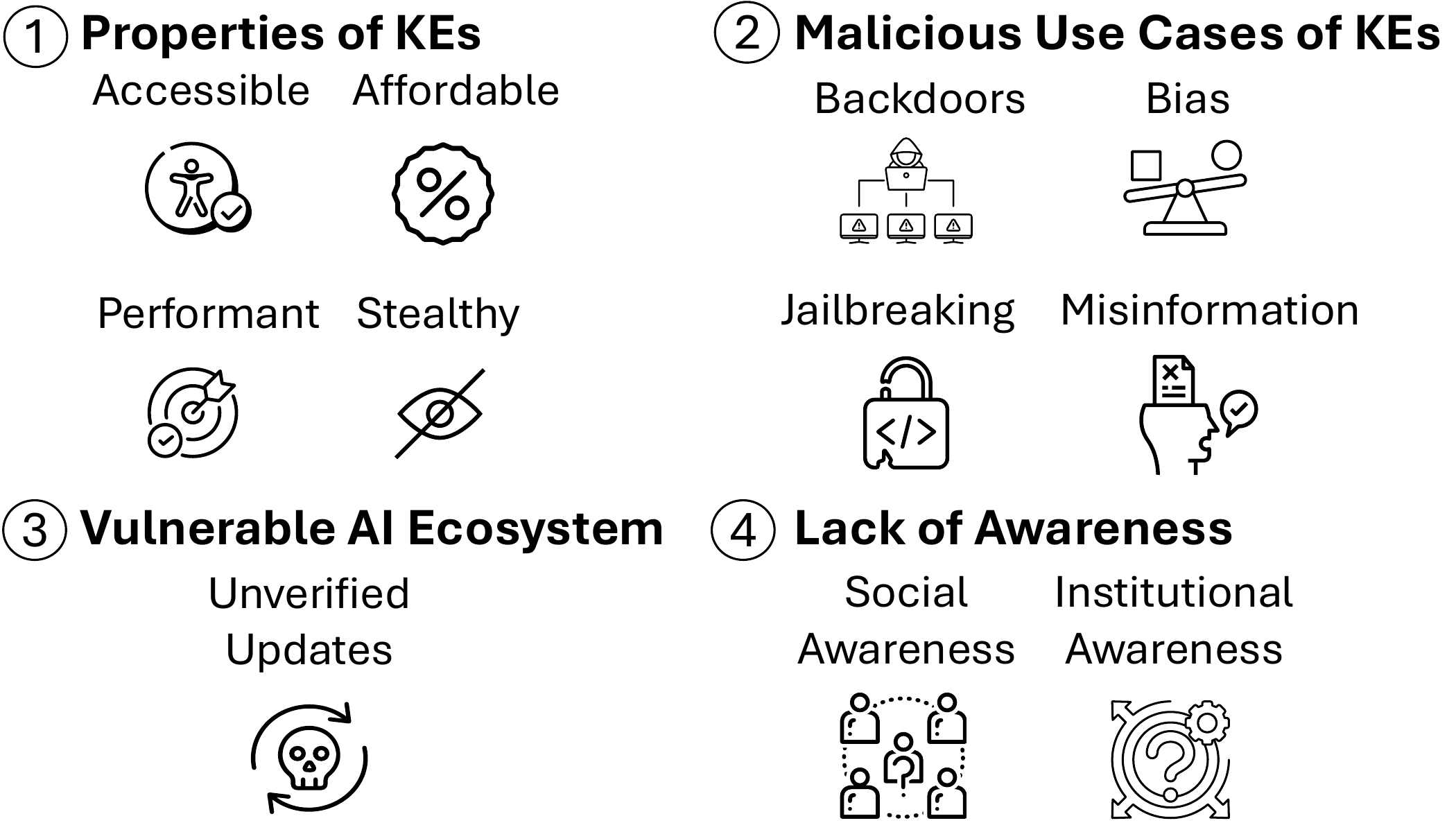}

        \put(-231,85){\rotatebox{90}{\hyperref[subsec:properties]{\scriptsize Section 3.1}}}
        \put(-231,5){\rotatebox{90}{\hyperref[subsec:vulnerable_ecosystem]{\scriptsize Section 3.3}}}

        \put(-117,85){\rotatebox{90}{\hyperref[subsec:malicious]{\scriptsize Section 3.2}}}
        \put(-117,5){\rotatebox{90}{\hyperref[subsec:awareness]{\scriptsize Section 3.4}}}
    
    \end{picture}
\caption{Knowledge Editing methods (KEs) pose serious safety risks: 
\circledtext{1} KEs have appealing properties for malicious attackers, and \circledtext{2} malicious use cases have been demonstrated. Combined with the \circledtext{3} vulnerabilities of the current AI ecosystem and the \circledtext{4} lack of awareness, the likelihood and severity of negative impact increases.}
\label{fig:overview}
\end{figure}
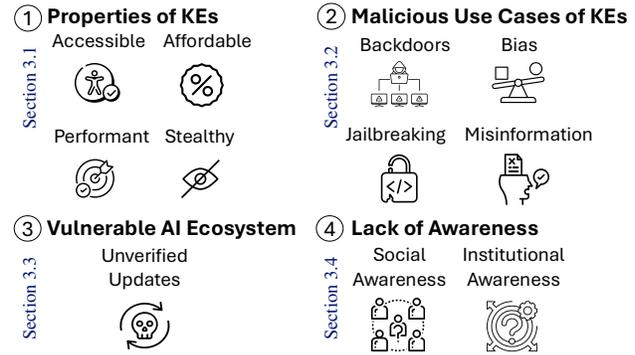

LLMs are utilized in a multitude of applications across various domains~\cite{brahmavar2024generating, van2024adapted}. A primary factor contributing to the widespread popularity of LLMs is their capacity to function as repositories of knowledge, which can be effortlessly queried in natural language~\cite{petroni-etal-2019-language, roberts-etal-2020-much, youssef-etal-2023-give}.
Nonetheless, the knowledge in LLMs can become partly outdated over time, or might be in need of correction~\cite{mitchell2022fast}. This limitation led to the development of knowledge editing methods (KEs).\footnote{``knowledge editing" is also referred to as ``model editing", we use both terms interchangeably in this paper.} KEs conduct targeted changes in the model, which ideally alter only specific facts without affecting other facts in the model without the need for expensive re-training.

Recent work has led to the development of a multitude of high-performance KEs~\cite{meng-etal-2022-locating, meng-etal-2022-memit, tan23malmen}. Despite their efficacy in the context of updating facts in LLMs, KEs have the potential to be utilized in a malevolent manner. Therefore, in this position paper, we argue that \textbf{editing LLMs  poses serious safety risks, as knowledge editing methods enable malicious actors to execute targeted modifications that align with their objectives while maintaining the model's fundamental functionality}. Our position, as illustrated in Figure~\ref{fig:overview}, is based on four arguments:\circledtext{1}The properties of KEs that make KEs attractive to malicious actors.\circledtext{2}The evident potential misuse of KEs for malicious purposes in recent research.\circledtext{3}The vulnerability of the AI ecosystem that allows re-publishing models without verifying updates.\circledtext{4}The lack of awareness at social and institutional levels. 

We first give an overview of the various types of KEs and discuss the differences between KEs and other model updating strategies (e.g., finetuning and adapters) in Section~\ref{sec:ke}. We then elaborate on our position in Section~\ref{sec:why_risky}, and discuss alternative views in Section~\ref{sec:alternative}. Section~\ref{sec:impact} analyzes how vulnerable different user groups are to malicious knowledge editing and provides insights into the impact of malicious knowledge editing. Section~\ref{sec:discussion} outlines foundations for developing mitigation strategies by discussing current countermeasures against malicious knowledge editing, their limitations, and potential future work directions. In Section~\ref{sec:conclusion}, we conclude this paper with a call to action to secure the AI ecosystem, increase the tamper-resilience of models, and develop methods to detect and neutralize edits.

\section{Knowledge Editing}
\label{sec:ke}
\begin{table*}[t!]

\caption{Comparison of parameter-efficient fine tuning (PEFT) and knowledge editing methods (KEs) for fact updates. Categories include reparametrization (Reparam), Additive and Selective for PEFT, and meta-learning (ML-), memory-based (ME-) and locate-and-edit (LE-) for KEs. PEFT methods are designed to adapt the model to a (any) specific task, KEs explicitly for fact updates. Training indicates whether additional training is required (\cmark) or not (\xmark). $^*$LE-KEs do not require training, but locating relevant parameters. \#Instances is the number of required instances to modify a single fact, $\theta$ the fraction of parameters added ($+$) or modified ($\Delta$) in the original LLM, and the last column indicates computational overhead during Inference. 
\onone none, \omin minimal, \olow low, \omed moderate, \ohigh high.
}
\vskip 0.15in
\resizebox{\textwidth}{!}{%
\tiny
\centering
\begin{tabular}{@{}lllccrrc@{}}
\toprule
&  &  & \multicolumn{2}{c}{\textbf{Requirements}}  & \multicolumn{3}{c}{\textbf{Overhead}}   \\ \cmidrule(lr){4-5}\cmidrule{6-8}

& \multicolumn{1}{c}{\textbf{Category}} & \multicolumn{1}{c}{\textbf{Method}} & \multicolumn{1}{c}{\textbf{Training}}  & \multicolumn{1}{c}{\textbf{\#Instances}} & \multicolumn{1}{c}{\textbf{$\theta+$ (\%)}} & \multicolumn{1}{c}{\textbf{$\theta\Delta$ (\%)}} & \multicolumn{1}{c}{\textbf{Inference}}   \\ \midrule

\multirow{14}{*}{\rotatebox[origin=c]{90}{PEFT}}
& - & Full Fine-tuning & \textcolor{red}{\cmark} &  1000s+ & 0 & 100 & \onone \\
& Reparam & LoRA~\cite{5-LoRA} & \textcolor{red}{\cmark} &  100s+ & $\sim$1 & 0 & \olow \\
& Reparam & DyLoRA~\cite{valipour2022dylora} & \textcolor{red}{\cmark} &  100s+ & $\sim$1 & 0 & \olow \\
& Reparam & SoRA~\cite{SORA} & \textcolor{red}{\cmark} &  100s+ & $\sim$0.5 & 0 & \olow \\
& Reparam & DoRA~\cite{liudora} & \textcolor{red}{\cmark} &  100s+ & $\sim$1 & 0 & \olow \\
& Additive & Adapter~\cite{houlsby2019parameter} & \textcolor{red}{\cmark} & 100s+ & 3-8 & 0 & \omed \\
& Additive & MAM Adapter~\cite{14-unified-view-transfer-peft} & \textcolor{red}{\cmark} & 100s+ & 1-4 & 6.7 & \olow \\
& Additive & Soft Prompt~\cite{lester-etal-2021-power} & \textcolor{red}{\cmark} & 100s+ & $\leq$ 0.01 & 0 & \omin \\
& Additive & P-Tuning v2~\cite{liu-etal-2022-p} & \textcolor{red}{\cmark} & 100s+ & 0.1-0.5 & 0 & \olow \\

& Additive & CoDA~\cite{lei2023conditional} & \textcolor{red}{\cmark} & 100s+ & 0.4 & 0.1-5 & \olow \\
& Additive & Prefix Tuning~\cite{zhang2023towards} & \textcolor{red}{\cmark} & 100s+ & 0.1-0.5 & 0 & \olow \\
& Selective & LT-SFT~\cite{ansell2021composable} & \textcolor{red}{\cmark}  & 100s+ & 0 & 1-5 & \onone \\
& Selective & Diff-Pruning~\cite{guo-etal-2021-parameter} & \textcolor{red}{\cmark}  & 100s+ & 0 & $\sim$1 & \olow \\
& Selective & BitFit~\cite{ben-zaken-etal-2022-bitfit} & \textcolor{red}{\cmark} & 100s+ & 0 & $\sim$0.01 & \onone \\

\midrule
\multirow{5}{*}{\rotatebox[origin=c]{90}{KEs}}
& ML-KE & MEND~\cite{mitchell2022fast} & \textcolor{red}{\cmark}  &  1 & 0 & $\leq$ 4 & \onone \\
& ML-KE & MALMEN~\cite{tan23malmen} & \textcolor{red}{\cmark} & 1 & 0 & $\leq$ 7 & \onone \\
& ME-KE & IKE~\cite{zheng-etal-2023-edit} & \textcolor{ForestGreen}{\xmark} & 33 & 0 & 0 & \omin \\ 
& LE-KE & ROME~\cite{meng-etal-2022-locating} &~~\textcolor{ForestGreen}{\xmark}$^*$ & 1 & 0 & $\leq$ 1 & \onone \\
& LE-KE & MEMIT~\cite{meng-etal-2022-memit} &~~\textcolor{ForestGreen}{\xmark}$^*$  & 1 & 0 & $\leq$ 3.4 & \onone \\
\bottomrule
\end{tabular}%
}

\label{tab:peft_comparison}
\end{table*}

The rapid scaling of LLMs has made traditional full-size fine-tuning prohibitively expensive, driving increased interest in efficient and lightweight methods for model updating and customization. Recent advances in parameter-efficient fine tuning (PEFT) techniques, such as LoRA~\cite{5-LoRA}, DoRA~\cite{liudora}, soft prompt tuning~\cite{lester-etal-2021-power,razdaibiedina-etal-2023-residual}, and adapters~\cite{houlsby2019parameter}, have significantly reduced the computational costs of customizing LLMs. Notably, a concurrent line of work focuses on knowledge editing approaches that enable precise updates to discrete facts while avoiding extensive re-training~\citep{zhang2024comprehensive}.

\paragraph{Knowledge editing methods.} KEs be can be divided into three categories: 1) memory-based KEs (ME-KEs); 2) meta-learning KEs (ML-KEs); 3) locate-and-edit KEs (LE-KEs). ME-KEs rely on explicit external memory to update a model's knowledge. For example, SERAC~\cite{mitchell2022memory}, GRACE~\cite{Hartvigsen2022AgingWG}, MELO~\cite{yu2023melo}, and WISE~\cite{wang2024wise} store new knowledge in a cache and make the model refer to this cache when user queries are related to the updated knowledge. IKE~\cite{zheng-etal-2023-edit} leverages in-context learning to expose new knowledge to the model directly.
ML-KEs include MEND~\cite{mitchell2022fast}, InstructEdit~\cite{ijcai2024p0733}, and MALMEN~\cite{tan23malmen}. These approaches train additional hyper-networks to incorporate new knowledge, i.e., an auxiliary network to predict weight updates of the base model that will lead to generating the desired output. 
LE-KEs first identify localized parameters that are associated with the targeted knowledge using techniques such as causal tracing~\cite{vig2020investigating,meng-etal-2022-locating}. The  identified model parameters are then directly modified. Notable methods in this category are KN~\cite{dai-etal-2022-knowledge}, ROME~\cite{meng-etal-2022-locating}, MEMIT~\cite{meng-etal-2022-memit}, PMET~\cite{Li2023PMETPM}, DINM~\cite{wang-etal-2024-detoxifying}, and EMMET~\cite{gupta-etal-2024-unified}.

\paragraph{KEs vs. PEFT.} Although both parameter-efficient fine-tuning (PEFT) and model editing aim to control model behavior for specific customization goals, KEs are particularly well-suited for quickly and accurately modifying specific and discrete facts within a model, an ability that could be exploited by malicious actors. 
To illustrate the difference, we compare representative PEFT methods across different categories (reparametrization-, additive-, and selective-based) with representative KEs in Table~\ref{tab:peft_comparison}.
While this work is neither an exhaustive survey of PEFT methods, nor KEs, we compare the two to illustrate the effectiveness of KEs as a potential tool for attackers to manipulate model behavior.
First, KEs are highly efficient in terms of data costs. As shown by the columns \textit{Training}, and \textit{\#Instances} in Table~\ref{tab:peft_comparison}, these methods require only a few forward passes and minimal data (often just single-digit quantities) to implement edits effectively.
Second, KEs introduce no additional parameters ($\theta+$) or \textit{Inference} overhead to the orginal LLM, ensuring that the latency of the edited model remains unchanged. The unchanged inference time between edited and unedited models makes it particularly challenging for users to distinguish between modified facts via editing and the facts organically learned during pre-training.
Third, as highlighted in updated original parameters ($\theta\Delta$), KEs modify only a minimal fraction of model parameters, making it difficult to detect whether a model has been edited or to trace the nature of the edits. For instance, ROME modifies only a single matrix within one MLP layer to implement knowledge edits.
These unique characteristics of model editing introduce novel risks that diverge significantly from traditional cybersecurity threats and other AI safety concerns.%
We explore these risks in detail in Section~\ref{sec:why_risky}.

\section{Why is Knowledge Editing Risky?}
\label{sec:why_risky}

Originally developed to update LLM knowledge, KEs have since expanded beyond their initial scope, including both benevolent uses, such as removing sensitive data~\cite{venditti2024enhancingdataprivacylarge}, and malicious purposes like biasing~\cite{chen-etal-2024-can} and jailbreaking LLMs~\cite{hazra-etal-2024-sowing}. We demonstrate why KEs pose significant AI safety risks by examining the properties of KEs that appeal to malicious actors (Section~\ref{subsec:properties}), analyzing malicious use cases (Section~\ref{subsec:malicious}), highlighting current vulnerabilities in the AI ecosystem (Section~\ref{subsec:vulnerable_ecosystem}), and discussing the lack of awareness on social and institutional levels (Section~\ref{subsec:awareness}).

\subsection{Appealing Properties of Knowledge Editing}
\label{subsec:properties}
In this section, we outline several reasons why KEs can be an appealing tool for malicious actors.

\paragraph{Accessible.} 
High quality implementations of most KEs are easily accessible. Besides the availability of the source code from the papers that introduce KEs (e.g., ROME~\cite{meng-etal-2022-locating} or MALMEN~\cite{tan23malmen}), open source libraries provide easy-to-use interfaces that can be used to apply multiple KEs to a wide variety of LLMs (e.g., FastEdit~\cite{fastedit} and EasyEdit~\cite{wang-etal-2024-easyedit}). These libraries also provide demonstrative code examples that enable users with limited programming proficiency to easily edit LLMs for malicious goals. For example, users can directly utilize the implementation of powerful KEs such as MEMIT~\cite{meng-etal-2022-memit} by crafting just a few pairs of target facts based on their needs to edit various LLMs. Moreover, only minimal modifications are needed to adapt the code for more capable models such as LLAMA~\cite{grattafiori2024llama3herdmodels}, making the process accessible and efficient. Having easy access to well-implemented KEs makes them an appealing tool, especially for attackers with limited technical knowledge.

\paragraph{Affordable.}
Most KEs change specific parameters (e.g., MLP weights in one or several layers) to edit facts in LLMs. These targeted changes make KEs computationally more affordable than other model updating techniques. Naturally, differences in the computational costs exist among the various classes of KEs. Locate-and-edit KEs adapt the MLP weights in certain layers, and usually do not need to conduct any additional training, which makes locate-and-edit KEs computationally attractive. Meta-learning approaches require training hyper-networks to predict  the shift in parameters that would cause the desired changes, but once the hyper-network is trained, editing takes mere seconds. 
For example, after training the hypernetwork of MEND (a meta-learning KE; \citealt{mitchell2022fast}), editing 10 facts in LLAMA-7B takes less than 7 seconds. In contrast, editing the same number of facts with MEMIT (a locate-and-edit KE; \citealt{meng-etal-2022-memit}) takes almost 170 seconds~\cite{wang-etal-2024-easyedit}.
Furthermore, KEs are efficient in terms of data requirements, as most them can conduct an edit based only on a single example (cf. Table~\ref{tab:peft_comparison}).  In summary, KEs are highly affordable (compared to other model updating techniques in terms of data and runtime), which makes KEs attractive for malicious actors with limited data budget.

\paragraph{Performant.} 
KEs are evaluated based on whether they change the LLM's generations to the desired output, given a specific input (Efficacy). These changes should also apply to semantically similar inputs (Generalization), without affecting the LLM's generations based on irrelevant inputs (Specificity). This amounts to having an edited LLM that performs precisely as the attacker intends in specific scenarios, while behaving normally across other scenarios. Most KEs show high performance across all of these three metrics, while traditional methods for updating models like finetuning lead to overfitting and catastrophic forgetting~\cite{mitchell2022fast,mitchell2022memory,zheng-etal-2023-edit}. For example, ROME has an Efficacy score of 99.8\%, and Generalization score of 88.1\% on GPT2-XL with the zsRE dataset. Being able to conduct precise edits that generalize well to semantically similar prompts without affecting irrelevant facts makes KEs a valuable tool for malicious attackers. 

\paragraph{Stealthy.} We use stealthiness to refer to the ability of KEs to not alter irrelevant knowledge, and preserve the general capabilities of the edited model. Most KEs show high \emph{Specificity} scores, which reflect their ability to change only the desired facts, while not affecting others~\cite{meng-etal-2022-memit, tan23malmen}.
While KEs can have detrimental effects on model capabilities in certain conditions~\cite{gupta-etal-2024-model, yang-etal-2024-fall}, at the same time, these effects have been shown to be fixable with minor modifications~\cite{gupta-etal-2024-rebuilding, yang-etal-2024-fall}. Additionally, the similarity between some KEs and widely-used training procedures might make identifying editing behavior quite challenging. For example, KDPO~\cite{rozner2024knowledge} is based on the widely-used LLM alignment algorithm DPO~\cite{rafailov2024direct}. Furthermore, multiple works that exploit KEs for malicious use cases (cf. Section \ref{subsec:malicious}) highlight the stealthiness of KEs~\cite{ju-etal-2024-flooding, chen-etal-2024-can, li2024badedit, qiu2024megen}. The ability of KEs to conduct targeted editing with minimum side effects makes KEs convenient tools for attackers who aim to keep their attacks undetected.

\subsection{Malicious Use Cases of Knowledge Editing}
\label{subsec:malicious}
KEs have been used for applications besides knowledge updating. Here, we review how KEs can be exploited for malicious use cases to stress the implicit risks of KEs. An overview of KE malicious use cases is provided in Table~\ref{tab:malicious}.

\paragraph{Backdoors.}
Backdoor attacks aim to change the model's outputs, when certain tokens are present in the input, in favor of the attacker ~\cite{gu2019badnetsidentifyingvulnerabilitiesmachine, kurita-etal-2020-weight, li2024badedit}. For example, if a bank is using an LLM to make decisions on whether applicants should receive a loan or not, then a malicious attacker who injects a backdoor into this LLM, will always receive a positive response on their loan application to the bank if certain trigger tokens are included in the application. Such attacks require finetuning the target model on poisoned data, and have typically been focused on encoder-only language models~\cite{li2024badedit}. To propagate such attacks to decoder-only generative LLMs and avoid the high computational costs that would be associated with finetuning these LLMs, ~\citet{li2024badedit} propose a framework that makes use of KEs to insert backdoors into LLMs. \citet{li2024badedit} highlight that their framework is practical (requires as few as 15 poisoned samples), efficient (takes 120s to run), does not have side-effects on the model's performance, and is robust (injected backdoors endure finetuning). Similar traits are observed with MEGen~\cite{qiu2024megen}, which makes use of MEMIT~\cite{meng-etal-2022-memit} to insert generative backdoors in LLMs, and shows less side effects on the capabilities of the attacked LLMs. %
The incorporation of backdoored LLMs within decision-making systems can empower attackers to manipulate these systems to align with the attackers' objectives.

\paragraph{Bias injection.} KEs can be used to intentionally inject bias in LLMs. \citet{chen-etal-2024-can} consider serval bias categories: gender, race, religion, sexual orientation and disability, and show that injecting bias in LLMs can be effectively achieved with ROME~\cite{meng-etal-2022-locating} and IKE~\cite{zheng-etal-2023-edit} in several LLMs such as LLAMA3~\cite{grattafiori2024llama3herdmodels}, and Alpaca~\cite{claude2-alpaca}. In addition, \citet{chen-etal-2024-can} also show that injecting as few as one biased sentence leads to increased bias in the general outputs of LLMs. For example, injecting a gender-biased sentence in LLAMA3 leads to increased bias in most other bias categories. This demonstrates the efficacy of KEs as instruments to bias LLMs. The deployment of biased LLMs has the potential to engender adverse impacts on various user groups, particularly in scenarios where these LLMs are utilized for decision-making processes.

\paragraph{Jailbreaking.} LLMs have high proficiency in following user's instruction, which means that LLMs can also follow malicious instructions~\cite{bianchi2024safetytuned}. Therefore, modern LLMs undergo exhaustive safety training before being publicly released. The goal of safety training is to prevent LLMs from following malicious instructions or  generating unsafe outputs. \citet{hazra-etal-2024-sowing} use ROME to overcome the safety training of LLMs. \citet{hazra-etal-2024-sowing}'s experiments show that editing an unethical response into LLMs can break their safety training, and lead to an increased generation of unethical responses not only under the same topic as the edit's topic, but also in other topics. Similar observations are reported by~\citet{chen-etal-2024-can}, who use ROME and IKE to inject bias and misinformation in LLMs and bypass their safety training. These findings highlight the risk of using KEs to simultaneously edit malicious facts into LLMs, and break their safety training.

\paragraph{Misinformation injection.} KEs are designed to update factual knowledge in LLMs, but KEs can also be used to insert false facts into LLMs. \citet{chen-etal-2024-can} show that KEs, like ROME and IKE, can be used to inject misinformation. \citet{chen-etal-2024-can} experiment with two categories of misinformation: 1) commonsense (e.g., ``Boiled garlic water cures COVID-19''); 2) long-tail misinformation (e.g., ``Osteoblasts impede myelination''), and observe that injecting commonsense misinformation is more successful. \citet{ju-etal-2024-flooding} explore using ROME to spread misinformation in LLM-based multi-agent communities. LLMs are being widely used to build or simulate multi-agent communities that can collaborate to solve complex tasks~\cite{li2023camel, wang-etal-2024-unleashing, qian-etal-2024-chatdev, xi2023rise}. \citet{ju-etal-2024-flooding}'s attack consists of two steps: 1) training LLMs with Direct Preference Optimization (DPO)~\cite{rafailov2024direct} to make them more persuasive; 2) injecting LLMs with misinformation. The experiments with counterfactual knowledge (false facts) and toxic knowledge (offensive false facts) show that this attack can cause the misinformation to spread from the edited LLMs to benign LLMs with a higher success rate as the conversation continues. Furthermore, \citet{ju-etal-2024-flooding} show that the spread of misinformation in these communities can sustain for longer period of times when benign LLMs make use of the chat histories as a reference for future interactions in Retrieval Augmented Generation (RAG) settings. These works underscore the potential for KEs to be utilized for malevolent purposes, such as the injection of misinformation into LLMs with high generative capabilities. Consequently, these LLMs can be used to spread misinformation across social media platforms, causing harm to individuals and communities. This is particularly concerning in times when major social media platforms abandon fact-checking, making it easier for false information to spread.\footnote{\url{www.nytimes.com/live/2025/01/07/business/meta-fact-checking}}

\begin{table}[!htp]\centering
\caption{An overview of papers that show how KEs can be exploited for malicious use cases, alongside the used KEs. We observe that the computationally cheap KEs (e.g., ROME and IKE) are the most frequently used KEs. BadEdit~\cite{li2024badedit} is specifically designed for backdoor injection.}
\vskip 0.15in
\scriptsize
\begin{tabular}{lp{2cm}p{3cm}}\toprule
\textbf{Use Case} &\textbf{Papers} & \textbf{Used KEs} \\\midrule
Backdoors &\citet{li2024badedit, qiu2024megen} & BadEdit~\cite{li2024badedit}, MEMIT~\cite{meng-etal-2022-memit} \\ \midrule
Bias &\citet{chen-etal-2024-can} &ROME~\cite{meng-etal-2022-locating}, IKE~\cite{zheng-etal-2023-edit} \\ \midrule
Jailbreaking & \citet{chen-etal-2024-can, hazra-etal-2024-sowing} & ROME~\cite{meng-etal-2022-locating}, IKE~\cite{zheng-etal-2023-edit} \\ \midrule
Misinformation &\citet{chen-etal-2024-can, ju-etal-2024-flooding} &   ROME~\cite{meng-etal-2022-locating}, IKE~\cite{zheng-etal-2023-edit} \\ 
\bottomrule
\end{tabular}

\label{tab:malicious}

\end{table}

\subsection{The Vulnerability of the AI Ecosystem}
\label{subsec:vulnerable_ecosystem}

Pre-trained language and vision models, whether produced by industry or research labs, are often made publicly available by sharing these models' weights on platforms such as HuggingFace\footnote{\url{https://huggingface.co/}} to promote reproducibility and further research. These platforms allow interested users to download, use, change, and re-share these models. Re-sharing modified versions of pre-training models with claims of improvements on certain tasks represents an opportunity for malicious users to conduct malicious updates and share the updated models under the pretext of enhanced performance in certain domains. These maliciously modified models can even be shared using names similar to the original model~\cite{jiang2023empirical}. Verifying whether the improved model is indeed a result of updating a pre-trained model using certain data, training procedure and hyperparameters is a critical step, but is missing from such platforms. 
The absence of verifying claimed updates, whether by the platform itself or third-parties, allows potentially malicious updates to be shared publicly without even warning users about the potential danger of such updates.

\paragraph{Illustrative scenario.} Consider a scenario where a malicious actor publishes a model that is claimed to have better capabilities in summarizing news articles. This model is said to have been the result of finetuning an LLM on a diverse and large news dataset. Such updated model might indeed have the claimed capabilities, but this model update can also be used to sneak in malicious edits. Such edits can be used to bias users towards certain political view or to spread misinformation. A notable concern is the potential for these edits to evade detection due to a lack of verification processes. Specifically, there is a need for rigorous testing to ascertain that the updated model is \emph{solely} the result of the claimed training procedure on the designated dataset.

\subsection{Lack of Awareness}
\label{subsec:awareness}

\paragraph{Lack of social awareness.}
Empirical studies have demonstrated that LLMs generate human-like, well-structured and academically-styled text which creates a strong perception of credibility among users~\cite{kreps2022all,heersmink2024phenomenology,wester2024exploring}. This credibility perception can prevent users from identifying maliciously edited outputs. Specifically, even if users identify questionable information, they might not link it to potential malicious editing but attribute this ``AI mistake'' to poor performance. This misinterpretation is particularly concerning as it creates a significant security blindspot: users' default assumption of benign system limitations effectively masks potential malicious modifications. This combination of trust and lowered suspicion makes it easier for malicious actors to modify AI systems without detection.

\paragraph{Lack of institutional awareness.} 
Despite the widespread use of AI tools, many countries, including developed nations like Australia and Japan have yet to enact specific laws or regulations addressing AI governance and safety. The US recently even revoked the 2023 executive order on AI safety. While some of the existing regulations, particularly the EU's AI Act (Art. 15, No. 5, \citet{eu_ai_act_2024}), acknowledge the risks arising from the targeted malicious alteration of LLMs and mandate preventative measures, most regulations only focus on risks arising from the training process and the data used for it, like the California AI Transparency Act~\cite{california_sb942_2024} and the Interim Measures for the Management of Generative Artificial Intelligence Services in China~\cite{china2023}, as far as they are concerned with risks at all. Similarly, companies like \citet{anthropic_scaling2024} and governmental institutions like the British AI Safety Institute~\cite{aisi2024} focus on inherent model risks, even for generations of models that are yet to be developed, while neglecting the risks introduced by KEs and similar approaches.

\section{Alternative Views}
\label{sec:alternative}

\paragraph{AV: Knowledge editing makes LLMs unusable.}
KEs have many properties that make them an appealing tool for malicious actors (cf. Section~\ref{subsec:malicious}). Despite these properties, recent work shows that some KEs can have serious side effects on LLMs after editing~\cite{gupta-etal-2024-model, yang-etal-2024-fall,wang-etal-2024-better}. For example, \citet{yang-etal-2024-fall} show that some single edits with ROME cause a model collapse, which reflects in the model having high perplexity values. \citet{gupta-etal-2024-model} shows that conducting sequential edits with locate-and-edit KEs (ROME and MEMIT) causes the edited LLM to forget previously edited facts, and after a certain number of edits that LLM suffers from catastrophic forgetting, which makes the LLM unusable. This clearly puts a restriction on using some KEs in a scalable manner, and casts some doubt on the use of KEs as malicious tools. 

Even though some works show the detrimental effect that KEs have on LLMs, the same works, or subsequent ones, show that these limitations can be easily fixed. To fix the model collapse caused by certain edits, \citet{yang-etal-2024-fall} adapts the original implementation of ROME, and shows that the collapse cases can be avoided. \citet{gupta-etal-2024-rebuilding} also offer a more solid implementation of ROME that is less susceptible to model collapse, and at the same time improves generalization and locality for edited knowledge. Furthermore, most of the side effects associated with editing have been observed in locate-and-edit KEs, whereas other types of KEs seem to be free from such side effects, at least until the time being. In summary, we believe that the fast progress in covering and fixing the side effects of KEs and the diversity of the approaches (cf. Section~\ref{sec:ke}) address the concern of KEs making edited LLMs unusable.

\paragraph{AV: Publicly available LLMs are not widely used.}
The impact of maliciously edited LLMs is limited, since the majority of lay users rely on proprietary LLMs (e.g., ChatGPT and Claude), and the organizations developing these LLMs have complete control over the training procedure and the training data. Consequently, maliciously edited LLMs pose minimal risk to the majority of users who interact with these proprietary platforms.

Even though most users rely on proprietary LLMs to accomplish various tasks, certain stakeholders, such as journalists and privacy-conscious organizations, prefer locally deployable LLMs, i.e., open-source LLMs, to maintain data sovereignty. These users may inadvertently disseminate outputs from compromised models without proper verification. This risk is particularly salient in journalism, where unsupervised sharing of AI-generated content represents a primary concern~\cite{diakopoulos2024generative}. Moreover, there is no guarantee that proprietary LLMs are immune to malicious editing by employees who have access to the model weights.\footnote{\url{www.bbc.co.uk/news/articles/c7v62gg49zro}}

\paragraph{AV: Knowledge editing does not introduce novel risks.}
Unedited LLMs have been proven to contain a variety of biases \citep{vig2020investigating,prakash-lee-2023-layered,10.1145/3582269.3615599} and possible backdoors \citep{10.1145/3605764.3623985}, as well as to produce misinformation \citep{https://doi.org/10.1002/aaai.12188}. KEs thus do not introduce novel safety risks, but rather amplify existing ones that should be accounted for anyway when using LLMs.

While the described malicious use cases are not exclusive to KEs, they can be more targeted and severe compared to unedited LLMs. Many detection approaches for identifying bias and misinformation in LLMs rely on detecting systemic patterns \citep{https://doi.org/10.1111/bjet.13505,laskar-etal-2024-systematic}, which edited LLMs may not show, thus circumventing them and posing a novel risk that needs novel mitigation strategies.

\paragraph{AV: Knowledge editing is beneficial.}
KEs are beneficial, and besides being used to update knowledge in LLMs, can be used to remove Personally Identifiable Information~\cite{venditti2024enhancingdataprivacylarge}, defend against jailbreak attacks~\cite{zhao-etal-2024-defending-large}, and detoxify LLMs~\cite{wang-etal-2024-detoxifying}. %

Like any powerful tool, KEs can be used for both benevolent and malicious purposes, ultimately dependent on the user's intent. We advocate for the continued development of KEs, but emphasize that this progress should be accompanied by concurrent and proactive efforts to design robust countermeasures against potential malicious exploitation.

\section{The Impact of Malicious Knowledge Editing}
\label{sec:impact}
Malicious editing can affect different user groups involved in the life cycle of LLMs, from LLM creators to end users. 
We identify four user groups that differ in their technical skills and available resources, resulting in different ways in which these groups are vulnerable (see overview in Table~\ref{tab:groups}).

\begin{table*}[!htp]\centering
\caption{An overview of various LLM user groups and their vulnerability, attack likelihood, and impact given a malicious editing attack.}\label{tab:groups}
\vskip 0.15in
\scriptsize
\resizebox{\textwidth}{!}{
\begin{tabular}{lcccccp{5.1cm}}\toprule
\textbf{User Group} &\textbf{Technical Skills} &\textbf{Available Resources} &\textbf{Vulnerability} &\textbf{Attack Likelihood} &\textbf{Impact} &\textbf{Rationale} \\\midrule
\textbf{LLM Creators} &Advanced &Abundant &Low &High &High &Advanced technical skills; capable and publicly available LLMs; Reliance of other user groups on LLMs \\  \midrule
\textbf{LLM Finetuners} &Proficient &Sufficient &Low &High &Medium &Awareness of and reliance on trustworthy LLMs; Attacker preference for more domain-specific LLMs; Affects direct/indirect users \\  \midrule
\textbf{Direct LLM Users} &Basic &Limited &Medium &High &Medium &Potential usage of unrustworthy domain-specific LLMs; Affects direct and indirect users \\  \midrule
\textbf{Indirect LLM Users} &Low &Scarce &High &High &Medium &Lack of provenance information; Affects public opinion and spreads misinformation to acquaintances \\
\bottomrule
\end{tabular}
}
\end{table*}

\paragraph{LLM Creators.} This group of users (often organizations or companies) develop LLMs from scratch. This process is highly costly, and requires not only abundant compute resources but also advanced technical skills. Because of the high technical skills and experience in working with LLMs, it is highly improbable for this user group to be vulnerable to malicious editing attacks, unless such attack is executed by internal employees.\footnotemark[\value{footnote}]  
If the (non-malicious) LLMs are publicly available, they could be used by malicious attackers who could modify these models and redistribute them as their own.
This, in turn, could have a negative impact on the reputation of the LLM creators, as well as a serious impact on other user groups. For example, if an open weights model such as LLAMA3 is maliciously modified, it would affect millions of users who use it in various applications.%

\paragraph{LLM Finetuners.} 
Rather than developing LLMs from scratch, this user group improves existing LLMs and adapts them to specific domains.
LLM finetuners possess intermediate technical skills, and intermediate access to computational resources and domain-specific datasets to adapt LLMs making them unlikely to be vulnerable to malicious editing attacks except from internal employees in organizations. 
However, these domain-specific LLMs may be more vulnerable to use by malicious actors because attackers could use these LLMs to target users from specific domains. Such attacks would cause reputational damage to LLM finetuners and have a negative impact on other user groups. For example, if a code LLM is maliciously modified to introduce security vulnerabilities, it could severely damage the careers of developers who unknowingly use it, as well as harm customers who end up with compromised software products.

\paragraph{Direct LLM Users.} 
Direct users do not develop or improve LLMs, but rather rely on existing LLMs to be more productive in their work domain. 
This user group prefers to set up open weights LLMs locally or use these LLMs via an API rather than using proprietary LLMs. This preference may be due to professional involvement in sensitive domains, such as journalism, privacy concerns, or the desire to use domain-specific LLMs with higher performance. Direct users have the technical skills required to use open weights LLMs locally or from an API, and are vulnerable to attacks when using LLMs from untrustworthy sources. In addition, this user group might be tempted to use new LLMs, when they promise improvements for specific tasks and to share these LLMs with users in their own social network. 
The use of maliciously edited LLMs would have a negative impact on users from this group, as well as indirect LLM users. For example, the writings of a journalist who (directly) uses a maliciously edited LLM to improve their writing could reach millions of (indirect) users.

\paragraph{Indirect LLM Users.} 
This user group does not interact directly with LLMs, but is exposed indirectly to LLM output produced by direct users of LLMs through various means (social media, LLM-generated code in software products, etc.). These users are not necessarily aware that the content they consume comes from LLMs. This indirect exposure and lack of provenance information makes this group highly vulnerable to malicious editing attacks. Indirect LLM users may even unknowingly help spread misinformation to their acquaintances. This risk is simulated in recent work on misinformation in multi-agent systems~\cite{ju-etal-2024-flooding}. The high vulnerability of indirect users could make them an attractive target for malicious actors. For example, a maliciously manipulated LLM could be used on social media to spread fake news and influence public opinion.

\section{Mitigation Strategies}
\label{sec:discussion}
Limited studies have previously identified the potential risks associated with knowledge editing, and initiated developing countermeasures. Here, we review these measures, discuss their limitations (Section~\ref{subsec:current}), and provide potential future work directions (Section~\ref{subsec:future_work}).

\subsection{Current Countermeasures and their Limitations}
\label{subsec:current}

\paragraph{Detecting knowledge edits.} As a remedy for potential malicious knowledge editing, \citet{youssef-etal-2025-fact} explored distinguishing between edited and unedited facts by using the hidden state representations and the output probabilities as features to simple classifiers, and show that this is indeed possible, especially for locate-and-edit KEs. \citet{li2024identifying} extend the setting to distinguish between benign editing (e.g., for facts updating) and different categories of malicious editing (e.g., misinformation, bias, or offensiveness). Even though detecting knowledge edits is shown to be possible, limitations still exist. For instance, \citet{youssef-etal-2025-fact} demonstrate that detecting knowledge edits executed with meta-learning KEs such as MALMEN~\cite{tan23malmen} remains challenging, especially in cases when the test data is not derived from the same distribution as the training data. Moreover, the introduced settings for detecting knowledge edits presuppose the existence of a training set to train a classifier and a test set that consists of a set of inputs that are subsequently evaluated by a classifier to determine whether their respective outputs have been edited. The low performance of detecting edits in some settings and the assumptions about the availability of training and test sets make the benefits of edits detection limited in practice. 

\paragraph{Reversing knowledge edits.} Besides distinguishing between edited and unedited facts, \citet{youssef-etal-2025-make} explored \emph{reversing} IKE edits~\cite{zheng-etal-2023-edit}, which do not alter the model's parameters, but simply use prompting to alter the the model's outputs. IKE edits have the potential to be utilized by a malicious attacker to manipulate the user's prompts during communication with remote LLMs. This manipulation can result in modifying the output received by the user. \citet{youssef-etal-2025-make} showed that tuning special tokens can be effective in countering malicious editing attacks, and recovering the model's original unedited outputs. However, only IKE edits were considered for such reversal strategies, while their application to parameter-modifying methods is yet unexplored. Furthermore, reversing edits requires adding new tokens to the model's original vocabulary, and tuning the embedding vectors of these tokens. Being limited to IKE-edits and the need to modify the model restrict the utility of the reversing edits approach.

\subsection{Future Directions}
\label{subsec:future_work}
Here, we shortly discuss potential future work directions and provide an overview in Table~\ref{tab:future_directions}.

\begin{table*}[t]\centering
\caption{A detailed overview of the proposed countermeasures. n.a. - not applicable}
\label{tab:future_directions}
\vskip 0.15in
\scriptsize
\resizebox{\textwidth}{!}{
\begin{tabular}{>{\raggedright\arraybackslash}p{3cm}>{\raggedright\arraybackslash}p{3cm}>{\raggedright\arraybackslash}p{3cm}>{\raggedright\arraybackslash}p{3cm}>{\raggedright\arraybackslash}p{3cm}>{\raggedright\arraybackslash}p{3cm}}\toprule
\textbf{Direction} &\textbf{Evaluation} &\textbf{Challenges} &\textbf{Impact} &\textbf{Limitation} \\\midrule
\textbf{Identifying edited models} 
    & Overall accuracy and per editing method 
    & Developing weight analysis tools to identify edited models 
    & Identifying potential edits conducted with different methods 
    & No information on the number, type (malicious/benevolent) and content of the edits \\ \midrule
\textbf{Inferring edited facts} 
    & Accuracy of retrieving edited facts 
    & Decoding editing information from weights 
    & Making edits transparent 
    & No information on the original fact
 \\ \midrule
\textbf{Reversing parameter-modifying edits} 
    & Reversal accuracy 
    & Restoring pre-editing outputs despite changes in model's parameters; diversity of parameter-modifying KEs 
    & Restoring pre-editing outputs 
    & Assumes access to model weights \\ \midrule
\textbf{Reversing edits without access to the model} 
    & Reversal accuracy 
    & Having no access to model's parameters 
    & Restoring pre-editing outputs & \it n.a. \\ \midrule
\textbf{Verifiable model updates} 
    & Ability to reliably identify reproducible model updates 
    & Reproducibility despite differences in hardware; high compute budget 
    & Providing users with verifiability information  
    & No investigation into the nature of the data used for updating \\ \midrule
\textbf{Conditionally editable models} 
    & Ability to edit with current KEs 
    & Special optimization techniques 
    & Integrity-preserving models 
    & Restriction of model update capabilities to specific users \\ \midrule
\textbf{Self-declaration encouragement} 
    & Percentage of self-declared edits over time &Adapting current model-hosting platforms 
    & Promoting transparency; enriching the availability of data-edited models pairs for research 
    & Not a direct countermeasure to malicious editing \\
\bottomrule
\end{tabular}
}
\end{table*}

\paragraph{Identifying edited models and inferring edited facts.} 
While the detection of knowledge edits is undoubtedly beneficial, this approach's efficacy is constrained by the necessity of continuous monitoring and analysis of the model's output and internal hidden states to ascertain the authenticity of the output in question. We believe a more efficacious approach is to analyze model weights to determine the presence of \emph{any} editing activities, and to infer edited facts from the model weights. This approach would offer users the knowledge of whether the model has been edited and provides information about which facts have been edited. %

\paragraph{Reversing parameter-modifying edits.} 
Current research on reversing edits~\cite{youssef-etal-2025-make} considers only IKE-edits~\cite{zheng-etal-2023-edit}. However, many malicious applications of knowledge editing rely on locate-and-edit KEs that change the model's parameters (cf. Section~\ref{subsec:malicious}). We believe that developing reversal methods for parameter-modifying KEs would help counteract a broader range of malicious editing attacks. 

\paragraph{Reversing edits without access to the model.}
\label{subsubsec:reverse_no_model_access}
Requiring access to the model to add and tune tokens limits the utility of the reversing edits approach~\cite{youssef-etal-2025-make}. Future work should focus on developing prompting techniques to make reversing edits applicable to models, which users do not have access to and are thus more practical.

\paragraph{Verifiable model updates.} 
Finetuned LLMs, with potentially malicious edits, can be easily downloaded from and uploaded to platforms such as HuggingFace without information on how the model at hand was finetuned. Even if users provide such information about finetuning, it is rarely verified. Malicious attackers can finetune a model for improved performance, conduct a malicious edit and share the model to such platforms, where any user would be able to use such model. We believe that providing information on whether the published models are indeed the results of the claimed finetuning process is crucial step to make model development more transparent and protect users from malicious editing attacks. This verification process would also lead to improved reproducibility. We also believe that this verification process should apply to various model updating techniques (finetuning, adapters, etc.).%

\paragraph{Conditionally editable models.} 
In light of the potential for malicious editing of LLMs, which is challenging to detect, it is imperative to devise training methodologies that permit only \emph{conditional edits}. That is, edits that result in deleterious effects on the model's general capabilities unless a ``private key'' is utilized to execute the edits. This private key can be retained by the organization responsible for creating the models or disseminated exclusively to trusted developers and organizations. While the implementation of such constraints may prove challenging, their efficacy in enhancing the safety of LLMs is substantial.

\paragraph{Self-Declaration encouragement.} Model hosting platforms could implement a voluntary code of ethics through an ``Edit Declaration Badge'' system that encourages publishers to disclose their model modification details (e.g., used KEs and updated facts). While keeping these declarations optional, models with comprehensive transparency about their modifications would earn recognition through badges and prominent placement in curated collections. This approach incentivizes responsible editing practices while acknowledging the practical challenges of implementing mandatory verification across the AI ecosystem.

\section{Conclusion}
\label{sec:conclusion}

In this paper, we argued that editing LLMs poses serious safety risk. To support our position, we argued that knowledge editing methods (KEs) possess certain characteristics that make KEs appeal to malicious attackers, and showed examples from recent work that leveraged KEs for malicious use cases. Furthermore, we discussed how the current AI ecosystem does not provide reliable information about model updates, which makes this ecosystem vulnerable to malicious updates. Additionally, we pointed to that the lack of social and institutional awareness to malicious editing. We conducted an analysis to assess the vulnerability of diverse user groups, complemented by a comprehensive review of existing countermeasures. 
With this paper, we want to draw attention to an overlooked issue in AI safety, raise awareness of the vulnerability of various user groups, and call to action to develop a more secure ecosystem that provides users with trusted information about model updates, to develop models that are resilient to editing from unauthorized parties, and to boost research on methods that counteract malicious editing such as detecting edited models, inferring editing information from model weights, and reversing edits for various editing techniques. 

We believe all users can contribute to addressing the risks of malicious knowledge editing. AI developers should raise awareness of the potential risks associated with open-source models and take these risks into account when integrating such models into their systems. Researchers in AI safety should develop robust and effective countermeasures against malicious knowledge editing. Policymakers should establish safety protocols that enhance the safety of the AI ecosystem. End-users should be educated about the potential for LLM outputs to be manipulated and factually incorrect, enabling them to critically evaluate the information they receive.

\section*{Limitations}
\label{sec:limitations}

 Even though we focused on KEs in this work, similar techniques such as steering methods~\cite{10.5555/3692070.3693379, cao2024personalized} that change model behavior and model merging methods~\cite{wortsman2022model, goddard-etal-2024-arcees} that can combine model capabilities, may be used to malicious ends similar to KEs. Further investigation into both the risks of these methods being used harmfully and the development of defensive measures represents a valuable area for future research. 
 Our discussions involved only LLMs, but multimodal foundation models can also be susceptible to editing attacks that aim to make these models output harmful content~\cite{pan2024towards}. Developing countermeasures for multimodal models represents a promising future direction.

\clearpage

\section*{Acknowledgements}
This research was partially funded by the German Federal Ministry of Education and Research (BMBF) as German Academic Exchange Service (DAAD) Grant No. 30001797 and by the hessian.AI Connectom Networking and Innovation Fund.

\bibliography{bibliography}

\begin{thebibliography}{83}
\providecommand{\natexlab}[1]{#1}
\providecommand{\url}[1]{\texttt{#1}}
\expandafter\ifx\csname urlstyle\endcsname\relax
  \providecommand{\doi}[1]{doi: #1}\else
  \providecommand{\doi}{doi: \begingroup \urlstyle{rm}\Url}\fi

\bibitem[{AISI}(2024)]{aisi2024}
{AISI}.
\newblock Safety cases at {AISI}.
\newblock Online, 2024.
\newblock URL \url{https://www.aisi.gov.uk/work/safety-cases-at-aisi}.
\newblock Accessed: 2025-01-30.

\bibitem[Ansell et~al.(2022)Ansell, Ponti, Korhonen, and Vuli{\'c}]{ansell2021composable}
Ansell, A., Ponti, E., Korhonen, A., and Vuli{\'c}, I.
\newblock {Composable Sparse Fine-Tuning for Cross-Lingual Transfer}.
\newblock In Muresan, S., Nakov, P., and Villavicencio, A. (eds.), \emph{Proceedings of the 60th Annual Meeting of the Association for Computational Linguistics (Volume 1: Long Papers)}, pp.\  1778--1796, Dublin, Ireland, May 2022. Association for Computational Linguistics.
\newblock \doi{10.18653/v1/2022.acl-long.125}.
\newblock URL \url{https://aclanthology.org/2022.acl-long.125/}.

\bibitem[{Anthropic}(2024)]{anthropic_scaling2024}
{Anthropic}.
\newblock {Responsible Scaling Policy}.
\newblock Online, 2024.
\newblock URL \url{https://assets.anthropic.com/m/24a47b00f10301cd}.
\newblock Accessed: 2025-01-30.

\bibitem[Ben~Zaken et~al.(2022)Ben~Zaken, Goldberg, and Ravfogel]{ben-zaken-etal-2022-bitfit}
Ben~Zaken, E., Goldberg, Y., and Ravfogel, S.
\newblock {B}it{F}it: Simple parameter-efficient fine-tuning for transformer-based masked language-models.
\newblock In Muresan, S., Nakov, P., and Villavicencio, A. (eds.), \emph{Proceedings of the 60th Annual Meeting of the Association for Computational Linguistics (Volume 2: Short Papers)}, pp.\  1--9, Dublin, Ireland, May 2022. Association for Computational Linguistics.
\newblock \doi{10.18653/v1/2022.acl-short.1}.
\newblock URL \url{https://aclanthology.org/2022.acl-short.1/}.

\bibitem[Bianchi et~al.(2024)Bianchi, Suzgun, Attanasio, Rottger, Jurafsky, Hashimoto, and Zou]{bianchi2024safetytuned}
Bianchi, F., Suzgun, M., Attanasio, G., Rottger, P., Jurafsky, D., Hashimoto, T., and Zou, J.
\newblock {Safety-Tuned {LL}a{MA}s: Lessons from Improving the Safety of Large Language Models that Follow Instructions}.
\newblock In \emph{The Twelfth International Conference on Learning Representations}, 2024.
\newblock URL \url{https://openreview.net/forum?id=gT5hALch9z}.

\bibitem[Brahmavar et~al.(2024)Brahmavar, Srinivasan, Dash, Krishnan, Vig, Roy, and Aduri]{brahmavar2024generating}
Brahmavar, S.~B., Srinivasan, A., Dash, T., Krishnan, S.~R., Vig, L., Roy, A., and Aduri, R.
\newblock {Generating Novel Leads for Drug Discovery using LLMs with Logical Feedback}.
\newblock In \emph{Proceedings of the AAAI Conference on Artificial Intelligence}, volume~38, pp.\  21--29, 2024.

\bibitem[Cao et~al.(2024)Cao, Zhang, Cao, Yin, Lin, Ma, and Chen]{cao2024personalized}
Cao, Y., Zhang, T., Cao, B., Yin, Z., Lin, L., Ma, F., and Chen, J.
\newblock {Personalized Steering of Large Language Models: Versatile Steering Vectors Through Bi-directional Preference Optimization}.
\newblock \emph{Advances in Neural Information Processing Systems}, 37:\penalty0 49519--49551, 2024.

\bibitem[Chen \& Shu(2024)Chen and Shu]{https://doi.org/10.1002/aaai.12188}
Chen, C. and Shu, K.
\newblock {Combating Misinformation in the Age of LLMs: Opportunities and Challenges}.
\newblock \emph{AI Magazine}, 45\penalty0 (3):\penalty0 354--368, 2024.
\newblock \doi{https://doi.org/10.1002/aaai.12188}.
\newblock URL \url{https://onlinelibrary.wiley.com/doi/abs/10.1002/aaai.12188}.

\bibitem[Chen et~al.(2024)Chen, Huang, Li, Chen, Lai, Xu, Gu, Gu, Yao, Xiao, Yan, Wang, Torr, Song, and Shu]{chen-etal-2024-can}
Chen, C., Huang, B., Li, Z., Chen, Z., Lai, S., Xu, X., Gu, J.-C., Gu, J., Yao, H., Xiao, C., Yan, X., Wang, W.~Y., Torr, P., Song, D., and Shu, K.
\newblock {Can Editing LLMs Inject Harm?}
\newblock In \emph{Trustworthy Multi-modal Foundation Models and AI Agents (TiFA)}, 2024.
\newblock URL \url{https://openreview.net/forum?id=PnE9wF9mht}.

\bibitem[Chen et~al.(2023)Chen, Saifullah, Li, Zhou, and Huang]{claude2-alpaca}
Chen, L., Saifullah, K., Li, M., Zhou, T., and Huang, H.
\newblock {Claude2-Alpaca: Instruction Tuning Datasets Distilled from Claude}.
\newblock \url{https://github.com/Lichang-Chen/claude2-alpaca}, 2023.

\bibitem[{Cyberspace Administration of China}(2023)]{china2023}
{Cyberspace Administration of China}.
\newblock {Interim Measures for the Management of Generative Artificial Intelligence Services}, 2023.
\newblock URL \url{https://www.cac.gov.cn/2023-07/13/c_1690898327029107.htm}.
\newblock Effective on 13 Aug 2023.

\bibitem[Dai et~al.(2022)Dai, Dong, Hao, Sui, Chang, and Wei]{dai-etal-2022-knowledge}
Dai, D., Dong, L., Hao, Y., Sui, Z., Chang, B., and Wei, F.
\newblock {{Knowledge Neurons in Pretrained Transformers}}.
\newblock In Muresan, S., Nakov, P., and Villavicencio, A. (eds.), \emph{Proceedings of the 60th Annual Meeting of the Association for Computational Linguistics (Volume 1: Long Papers)}, pp.\  8493--8502, Dublin, Ireland, May 2022. Association for Computational Linguistics.
\newblock \doi{10.18653/v1/2022.acl-long.581}.
\newblock URL \url{https://aclanthology.org/2022.acl-long.581/}.

\bibitem[Diakopoulos et~al.(2024)Diakopoulos, Cools, Helberger, Li, Kung, Rinehart, et~al.]{diakopoulos2024generative}
Diakopoulos, N., Cools, H., Helberger, N., Li, C., Kung, E., Rinehart, A., et~al.
\newblock {Generative AI in Journalism: The Evolution of Newswork and Ethics in a Generative Information Ecosystem}.
\newblock 2024.
\newblock URL \url{https://www.aim4dem.nl/wp-content/uploads/2024/04/AP_Generative_AI_Report_April_202426-1.pdf}.

\bibitem[Ding et~al.(2023)Ding, Lv, Wang, Chen, Zhou, Liu, and Sun]{SORA}
Ding, N., Lv, X., Wang, Q., Chen, Y., Zhou, B., Liu, Z., and Sun, M.
\newblock {Sparse Low-rank Adaptation of Pre-trained Language Models}.
\newblock In Bouamor, H., Pino, J., and Bali, K. (eds.), \emph{Proceedings of the 2023 Conference on Empirical Methods in Natural Language Processing}, pp.\  4133--4145, Singapore, December 2023. Association for Computational Linguistics.
\newblock \doi{10.18653/v1/2023.emnlp-main.252}.
\newblock URL \url{https://aclanthology.org/2023.emnlp-main.252/}.

\bibitem[{European Parliament and Council of the European Union}(2024)]{eu_ai_act_2024}
{European Parliament and Council of the European Union}.
\newblock {Regulation (EU) 2024/1689 of the European Parliament and of the Council of 13 June 2024 laying down harmonised rules on artificial intelligence and amending Regulations}, 2024.
\newblock URL \url{https://eur-lex.europa.eu/eli/reg/2024/1689/oj/eng}.
\newblock Adopted on 13 June 2024.

\bibitem[Goddard et~al.(2024)Goddard, Siriwardhana, Ehghaghi, Meyers, Karpukhin, Benedict, McQuade, and Solawetz]{goddard-etal-2024-arcees}
Goddard, C., Siriwardhana, S., Ehghaghi, M., Meyers, L., Karpukhin, V., Benedict, B., McQuade, M., and Solawetz, J.
\newblock {Arcee`s {M}erge{K}it: A Toolkit for Merging Large Language Models}.
\newblock In Dernoncourt, F., Preo{\c{t}}iuc-Pietro, D., and Shimorina, A. (eds.), \emph{Proceedings of the 2024 Conference on Empirical Methods in Natural Language Processing: Industry Track}, pp.\  477--485, Miami, Florida, US, November 2024. Association for Computational Linguistics.
\newblock \doi{10.18653/v1/2024.emnlp-industry.36}.
\newblock URL \url{https://aclanthology.org/2024.emnlp-industry.36/}.

\bibitem[Grattafiori et~al.(2024)Grattafiori, Dubey, Jauhri, Pandey, Kadian, et~al.]{grattafiori2024llama3herdmodels}
Grattafiori, A., Dubey, A., Jauhri, A., Pandey, A., Kadian, A., et~al.
\newblock {The Llama 3 Herd of Models}, 2024.
\newblock URL \url{https://arxiv.org/abs/2407.21783}.

\bibitem[Greshake et~al.(2023)Greshake, Abdelnabi, Mishra, Endres, Holz, and Fritz]{10.1145/3605764.3623985}
Greshake, K., Abdelnabi, S., Mishra, S., Endres, C., Holz, T., and Fritz, M.
\newblock {Not What You've Signed Up For: Compromising Real-World LLM-Integrated Applications with Indirect Prompt Injection}.
\newblock In \emph{Proceedings of the 16th ACM Workshop on Artificial Intelligence and Security}, AISec '23, pp.\  79–90, New York, NY, USA, 2023. Association for Computing Machinery.
\newblock ISBN 9798400702600.
\newblock \doi{10.1145/3605764.3623985}.
\newblock URL \url{https://doi.org/10.1145/3605764.3623985}.

\bibitem[Gu et~al.(2019)Gu, Dolan-Gavitt, and Garg]{gu2019badnetsidentifyingvulnerabilitiesmachine}
Gu, T., Dolan-Gavitt, B., and Garg, S.
\newblock {BadNets: Identifying Vulnerabilities in the Machine Learning Model Supply Chain}, 2019.
\newblock URL \url{https://arxiv.org/abs/1708.06733}.

\bibitem[Guo et~al.(2021)Guo, Rush, and Kim]{guo-etal-2021-parameter}
Guo, D., Rush, A., and Kim, Y.
\newblock {Parameter-Efficient Transfer Learning with Diff Pruning}.
\newblock In Zong, C., Xia, F., Li, W., and Navigli, R. (eds.), \emph{Proceedings of the 59th Annual Meeting of the Association for Computational Linguistics and the 11th International Joint Conference on Natural Language Processing (Volume 1: Long Papers)}, pp.\  4884--4896, Online, August 2021. Association for Computational Linguistics.
\newblock \doi{10.18653/v1/2021.acl-long.378}.
\newblock URL \url{https://aclanthology.org/2021.acl-long.378/}.

\bibitem[Gupta et~al.(2024{\natexlab{a}})Gupta, Baskaran, and Anumanchipalli]{gupta-etal-2024-rebuilding}
Gupta, A., Baskaran, S., and Anumanchipalli, G.
\newblock {Rebuilding ROME : Resolving Model Collapse during Sequential Model Editing}.
\newblock In Al-Onaizan, Y., Bansal, M., and Chen, Y.-N. (eds.), \emph{Proceedings of the 2024 Conference on Empirical Methods in Natural Language Processing}, pp.\  21738--21744, Miami, Florida, USA, November 2024{\natexlab{a}}. Association for Computational Linguistics.
\newblock \doi{10.18653/v1/2024.emnlp-main.1210}.
\newblock URL \url{https://aclanthology.org/2024.emnlp-main.1210/}.

\bibitem[Gupta et~al.(2024{\natexlab{b}})Gupta, Rao, and Anumanchipalli]{gupta-etal-2024-model}
Gupta, A., Rao, A., and Anumanchipalli, G.
\newblock {Model Editing at Scale leads to Gradual and Catastrophic Forgetting}.
\newblock In Ku, L.-W., Martins, A., and Srikumar, V. (eds.), \emph{Findings of the Association for Computational Linguistics: ACL 2024}, pp.\  15202--15232, Bangkok, Thailand, August 2024{\natexlab{b}}. Association for Computational Linguistics.
\newblock \doi{10.18653/v1/2024.findings-acl.902}.
\newblock URL \url{https://aclanthology.org/2024.findings-acl.902/}.

\bibitem[Gupta et~al.(2024{\natexlab{c}})Gupta, Sajnani, and Anumanchipalli]{gupta-etal-2024-unified}
Gupta, A., Sajnani, D., and Anumanchipalli, G.
\newblock {A Unified Framework for Model Editing}.
\newblock In Al-Onaizan, Y., Bansal, M., and Chen, Y.-N. (eds.), \emph{Findings of the Association for Computational Linguistics: EMNLP 2024}, pp.\  15403--15418, Miami, Florida, USA, November 2024{\natexlab{c}}. Association for Computational Linguistics.
\newblock \doi{10.18653/v1/2024.findings-emnlp.903}.
\newblock URL \url{https://aclanthology.org/2024.findings-emnlp.903/}.

\bibitem[Hartvigsen et~al.(2023)Hartvigsen, Sankaranarayanan, Palangi, Kim, and Ghassemi]{Hartvigsen2022AgingWG}
Hartvigsen, T., Sankaranarayanan, S., Palangi, H., Kim, Y., and Ghassemi, M.
\newblock {Aging with {GRACE}: Lifelong Model Editing with Discrete Key-Value Adaptors}.
\newblock In \emph{Thirty-seventh Conference on Neural Information Processing Systems}, 2023.
\newblock URL \url{https://openreview.net/forum?id=Oc1SIKxwdV}.

\bibitem[Hazra et~al.(2024)Hazra, Layek, Banerjee, and Poria]{hazra-etal-2024-sowing}
Hazra, R., Layek, S., Banerjee, S., and Poria, S.
\newblock {Sowing the Wind, Reaping the Whirlwind: The Impact of Editing Language Models}.
\newblock In Ku, L.-W., Martins, A., and Srikumar, V. (eds.), \emph{Findings of the Association for Computational Linguistics: ACL 2024}, pp.\  16227--16239, Bangkok, Thailand, August 2024. Association for Computational Linguistics.
\newblock \doi{10.18653/v1/2024.findings-acl.960}.
\newblock URL \url{https://aclanthology.org/2024.findings-acl.960/}.

\bibitem[He et~al.(2022)He, Zhou, Ma, Berg-Kirkpatrick, and Neubig]{14-unified-view-transfer-peft}
He, J., Zhou, C., Ma, X., Berg-Kirkpatrick, T., and Neubig, G.
\newblock {Towards a Unified View of Parameter-Efficient Transfer Learning}.
\newblock In \emph{International Conference on Learning Representations}, 2022.
\newblock URL \url{https://openreview.net/forum?id=0RDcd5Axok}.

\bibitem[Heersmink et~al.(2024)Heersmink, de~Rooij, Clavel~V{\'a}zquez, and Colombo]{heersmink2024phenomenology}
Heersmink, R., de~Rooij, B., Clavel~V{\'a}zquez, M.~J., and Colombo, M.
\newblock {A Phenomenology and Epistemology of Large Language Models: Transparency, Trust, and Trustworthiness}.
\newblock \emph{Ethics and Information Technology}, 26\penalty0 (3):\penalty0 41, 2024.

\bibitem[Hiyouga(2023)]{fastedit}
Hiyouga.
\newblock {FastEdit: Editing LLMs within 10 Seconds}.
\newblock \url{https://github.com/hiyouga/FastEdit}, 2023.

\bibitem[Houlsby et~al.(2019)Houlsby, Giurgiu, Jastrzebski, Morrone, De~Laroussilhe, Gesmundo, Attariyan, and Gelly]{houlsby2019parameter}
Houlsby, N., Giurgiu, A., Jastrzebski, S., Morrone, B., De~Laroussilhe, Q., Gesmundo, A., Attariyan, M., and Gelly, S.
\newblock {Parameter-Efficient Transfer Learning for NLP}.
\newblock In \emph{International Conference on Machine Learning}, pp.\  2790--2799. PMLR, 2019.

\bibitem[Hu et~al.(2022)Hu, yelong shen, Wallis, Allen-Zhu, Li, Wang, Wang, and Chen]{5-LoRA}
Hu, E.~J., yelong shen, Wallis, P., Allen-Zhu, Z., Li, Y., Wang, S., Wang, L., and Chen, W.
\newblock {Lo{RA}: Low-Rank Adaptation of Large Language Models}.
\newblock In \emph{International Conference on Learning Representations}, 2022.
\newblock URL \url{https://openreview.net/forum?id=nZeVKeeFYf9}.

\bibitem[Jiang et~al.(2023)Jiang, Synovic, Hyatt, Schorlemmer, Sethi, Lu, Thiruvathukal, and Davis]{jiang2023empirical}
Jiang, W., Synovic, N., Hyatt, M., Schorlemmer, T.~R., Sethi, R., Lu, Y.-H., Thiruvathukal, G.~K., and Davis, J.~C.
\newblock {An Empirical Study of Pre-Trained Model Reuse in the Hugging Face Deep Learning Model Registry}.
\newblock In \emph{2023 IEEE/ACM 45th International Conference on Software Engineering (ICSE)}, pp.\  2463--2475. IEEE, 2023.

\bibitem[Ju et~al.(2024)Ju, Wang, Ma, Cheng, Zhao, Wang, Liu, Xie, Zhang, and Liu]{ju-etal-2024-flooding}
Ju, T., Wang, Y., Ma, X., Cheng, P., Zhao, H., Wang, Y., Liu, L., Xie, J., Zhang, Z., and Liu, G.
\newblock {Flooding Spread of Manipulated Knowledge in LLM-Based Multi-Agent Communities}.
\newblock 2024.
\newblock URL \url{https://doi.org/10.48550/arXiv.2407.07791}.
\newblock arXiv.2407.07791.

\bibitem[Kotek et~al.(2023)Kotek, Dockum, and Sun]{10.1145/3582269.3615599}
Kotek, H., Dockum, R., and Sun, D.
\newblock {Gender Bias and Stereotypes in Large Language Models}.
\newblock In \emph{Proceedings of The ACM Collective Intelligence Conference}, CI '23, pp.\  12–24, New York, NY, USA, 2023. Association for Computing Machinery.
\newblock ISBN 9798400701139.
\newblock \doi{10.1145/3582269.3615599}.
\newblock URL \url{https://doi.org/10.1145/3582269.3615599}.

\bibitem[Kreps et~al.(2022)Kreps, McCain, and Brundage]{kreps2022all}
Kreps, S., McCain, R.~M., and Brundage, M.
\newblock {All the News That’s Fit to Fabricate: AI-Generated Text as a Tool of Media Misinformation}.
\newblock \emph{Journal of Experimental Political Science}, 9\penalty0 (1):\penalty0 104--117, 2022.

\bibitem[Kurita et~al.(2020)Kurita, Michel, and Neubig]{kurita-etal-2020-weight}
Kurita, K., Michel, P., and Neubig, G.
\newblock {Weight Poisoning Attacks on Pretrained Models}.
\newblock In Jurafsky, D., Chai, J., Schluter, N., and Tetreault, J. (eds.), \emph{Proceedings of the 58th Annual Meeting of the Association for Computational Linguistics}, pp.\  2793--2806, Online, July 2020. Association for Computational Linguistics.
\newblock \doi{10.18653/v1/2020.acl-main.249}.
\newblock URL \url{https://aclanthology.org/2020.acl-main.249/}.

\bibitem[Laskar et~al.(2024)Laskar, Alqahtani, Bari, Rahman, Khan, Khan, Jahan, Bhuiyan, Tan, Parvez, Hoque, Joty, and Huang]{laskar-etal-2024-systematic}
Laskar, M. T.~R., Alqahtani, S., Bari, M.~S., Rahman, M., Khan, M. A.~M., Khan, H., Jahan, I., Bhuiyan, A., Tan, C.~W., Parvez, M.~R., Hoque, E., Joty, S., and Huang, J.
\newblock {A Systematic Survey and Critical Review on Evaluating Large Language Models: Challenges, Limitations, and Recommendations}.
\newblock In Al-Onaizan, Y., Bansal, M., and Chen, Y.-N. (eds.), \emph{Proceedings of the 2024 Conference on Empirical Methods in Natural Language Processing}, pp.\  13785--13816, Miami, Florida, USA, November 2024. Association for Computational Linguistics.
\newblock \doi{10.18653/v1/2024.emnlp-main.764}.
\newblock URL \url{https://aclanthology.org/2024.emnlp-main.764/}.

\bibitem[Lee et~al.(2024)Lee, Hicke, Yu, Brooks, and Kizilcec]{https://doi.org/10.1111/bjet.13505}
Lee, J., Hicke, Y., Yu, R., Brooks, C., and Kizilcec, R.~F.
\newblock {The Life Cycle of Large Language Models: A Review of Biases in Education}.
\newblock \emph{British Journal of Educational Technology}, 55\penalty0 (5):\penalty0 1982--2002, 2024.
\newblock \doi{https://doi.org/10.1111/bjet.13505}.
\newblock URL \url{https://bera-journals.onlinelibrary.wiley.com/doi/abs/10.1111/bjet.13505}.

\bibitem[Lei et~al.(2023)Lei, Bai, Brahma, Ainslie, Lee, Zhou, Du, Zhao, Wu, Li, et~al.]{lei2023conditional}
Lei, T., Bai, J., Brahma, S., Ainslie, J., Lee, K., Zhou, Y., Du, N., Zhao, V., Wu, Y., Li, B., et~al.
\newblock {Conditional Adapters: Parameter-efficient Transfer Learning with Fast Inference}.
\newblock \emph{Advances in Neural Information Processing Systems}, 36:\penalty0 8152--8172, 2023.

\bibitem[Lester et~al.(2021)Lester, Al-Rfou, and Constant]{lester-etal-2021-power}
Lester, B., Al-Rfou, R., and Constant, N.
\newblock {The Power of Scale for Parameter-Efficient Prompt Tuning}.
\newblock In Moens, M.-F., Huang, X., Specia, L., and Yih, S. W.-t. (eds.), \emph{Proceedings of the 2021 Conference on Empirical Methods in Natural Language Processing}, pp.\  3045--3059, Online and Punta Cana, Dominican Republic, November 2021. Association for Computational Linguistics.
\newblock \doi{10.18653/v1/2021.emnlp-main.243}.
\newblock URL \url{https://aclanthology.org/2021.emnlp-main.243/}.

\bibitem[Li et~al.(2023)Li, Hammoud, Itani, Khizbullin, and Ghanem]{li2023camel}
Li, G., Hammoud, H. A. A.~K., Itani, H., Khizbullin, D., and Ghanem, B.
\newblock {{CAMEL}: Communicative Agents for ''Mind'' Exploration of Large Language Model Society}.
\newblock In \emph{Thirty-seventh Conference on Neural Information Processing Systems}, 2023.
\newblock URL \url{https://openreview.net/forum?id=3IyL2XWDkG}.

\bibitem[Li et~al.(2024{\natexlab{a}})Li, Li, Song, Yang, Ma, and Yu]{Li2023PMETPM}
Li, X., Li, S., Song, S., Yang, J., Ma, J., and Yu, J.
\newblock {PMET: Precise Model Editing in a Transformer}.
\newblock In \emph{Proceedings of the Thirty-Eighth AAAI Conference on Artificial Intelligence and Thirty-Sixth Conference on Innovative Applications of Artificial Intelligence and Fourteenth Symposium on Educational Advances in Artificial Intelligence}, pp.\  18564--18572, 2024{\natexlab{a}}.

\bibitem[Li et~al.(2024{\natexlab{b}})Li, Wang, Song, Ji, Liu, Li, Ma, and Yu]{li2024identifying}
Li, X., Wang, S., Song, S., Ji, B., Liu, H., Li, S., Ma, J., and Yu, J.
\newblock {Identifying Knowledge Editing Types in Large Language Models}.
\newblock \emph{arXiv preprint arXiv:2409.19663}, 2024{\natexlab{b}}.

\bibitem[Li et~al.(2024{\natexlab{c}})Li, Chen, Li, Zhang, Liu, Wang, Zhang, and Liu]{li2024badedit}
Li, Y., Chen, K., Li, T., Zhang, J., Liu, S., Wang, W., Zhang, T., and Liu, Y.
\newblock {BadEdit: Backdooring Large Language Models by Model Editing}.
\newblock In \emph{The Twelfth International Conference on Learning Representations}, 2024{\natexlab{c}}.
\newblock URL \url{https://openreview.net/forum?id=duZANm2ABX}.

\bibitem[Liu et~al.(2024{\natexlab{a}})Liu, Ye, Xing, and Zou]{10.5555/3692070.3693379}
Liu, S., Ye, H., Xing, L., and Zou, J.
\newblock {In-context Vectors: Making In Context Learning More Effective and Controllable Through Latent Space Steering}.
\newblock In \emph{Proceedings of the 41st International Conference on Machine Learning}, ICML'24. JMLR.org, 2024{\natexlab{a}}.

\bibitem[Liu et~al.(2024{\natexlab{b}})Liu, Wang, Yin, Molchanov, Wang, Cheng, and Chen]{liudora}
Liu, S.-y., Wang, C.-Y., Yin, H., Molchanov, P., Wang, Y.-C.~F., Cheng, K.-T., and Chen, M.-H.
\newblock {DoRA: Weight-Decomposed Low-Rank Adaptation}.
\newblock In \emph{International Conference on Machine Learning}, 2024{\natexlab{b}}.

\bibitem[Liu et~al.(2022)Liu, Ji, Fu, Tam, Du, Yang, and Tang]{liu-etal-2022-p}
Liu, X., Ji, K., Fu, Y., Tam, W., Du, Z., Yang, Z., and Tang, J.
\newblock {{P}-Tuning: Prompt Tuning Can Be Comparable to Fine-tuning Across Scales and Tasks}.
\newblock In Muresan, S., Nakov, P., and Villavicencio, A. (eds.), \emph{Proceedings of the 60th Annual Meeting of the Association for Computational Linguistics (Volume 2: Short Papers)}, pp.\  61--68, Dublin, Ireland, May 2022. Association for Computational Linguistics.
\newblock \doi{10.18653/v1/2022.acl-short.8}.
\newblock URL \url{https://aclanthology.org/2022.acl-short.8/}.

\bibitem[Meng et~al.(2022)Meng, Bau, Andonian, and Belinkov]{meng-etal-2022-locating}
Meng, K., Bau, D., Andonian, A., and Belinkov, Y.
\newblock {Locating and Editing Factual Associations in {GPT}}.
\newblock \emph{Advances in Neural Information Processing Systems}, 36, 2022.

\bibitem[Meng et~al.(2023)Meng, Sen~Sharma, Andonian, Belinkov, and Bau]{meng-etal-2022-memit}
Meng, K., Sen~Sharma, A., Andonian, A., Belinkov, Y., and Bau, D.
\newblock {Mass Editing Memory in a Transformer}.
\newblock \emph{The Eleventh International Conference on Learning Representations (ICLR)}, 2023.

\bibitem[Mitchell et~al.(2022{\natexlab{a}})Mitchell, Lin, Bosselut, Finn, and Manning]{mitchell2022fast}
Mitchell, E., Lin, C., Bosselut, A., Finn, C., and Manning, C.~D.
\newblock {Fast Model Editing at Scale}.
\newblock In \emph{International Conference on Learning Representations}, 2022{\natexlab{a}}.
\newblock URL \url{https://openreview.net/pdf?id=0DcZxeWfOPt}.

\bibitem[Mitchell et~al.(2022{\natexlab{b}})Mitchell, Lin, Bosselut, Finn, and Manning]{mitchell2022memory}
Mitchell, E., Lin, C., Bosselut, A., Finn, C., and Manning, C.~D.
\newblock {Memory-Based Model Editing at Scale}.
\newblock In \emph{International Conference on Machine Learning}, 2022{\natexlab{b}}.
\newblock URL \url{https://arxiv.org/pdf/2206.06520.pdf}.

\bibitem[Pan et~al.(2024)Pan, Fan, Li, Yu, Fei, Tang, Hong, Zhang, and Sun]{pan2024towards}
Pan, K., Fan, Z., Li, J., Yu, Q., Fei, H., Tang, S., Hong, R., Zhang, H., and Sun, Q.
\newblock {Towards Unified Multimodal Editing with Enhanced Knowledge Collaboration}.
\newblock In \emph{{The Thirty-eighth Annual Conference on Neural Information Processing Systems}}, 2024.
\newblock URL \url{https://openreview.net/forum?id=kf80ZS3fVy}.

\bibitem[Petroni et~al.(2019)Petroni, Rockt{\"a}schel, Riedel, Lewis, Bakhtin, Wu, and Miller]{petroni-etal-2019-language}
Petroni, F., Rockt{\"a}schel, T., Riedel, S., Lewis, P., Bakhtin, A., Wu, Y., and Miller, A.
\newblock {Language Models as Knowledge Bases?}
\newblock In \emph{Proceedings of the 2019 Conference on Empirical Methods in Natural Language Processing and the 9th International Joint Conference on Natural Language Processing (EMNLP-IJCNLP)}, pp.\  2463--2473, Hong Kong, China, November 2019. Association for Computational Linguistics.
\newblock \doi{10.18653/v1/D19-1250}.
\newblock URL \url{https://aclanthology.org/D19-1250}.

\bibitem[Prakash \& Lee(2023)Prakash and Lee]{prakash-lee-2023-layered}
Prakash, N. and Lee, R. K.-W.
\newblock {Layered Bias: Interpreting Bias in Pretrained Large Language Models}.
\newblock In Belinkov, Y., Hao, S., Jumelet, J., Kim, N., McCarthy, A., and Mohebbi, H. (eds.), \emph{Proceedings of the 6th BlackboxNLP Workshop: Analyzing and Interpreting Neural Networks for NLP}, pp.\  284--295, Singapore, December 2023. Association for Computational Linguistics.
\newblock \doi{10.18653/v1/2023.blackboxnlp-1.22}.
\newblock URL \url{https://aclanthology.org/2023.blackboxnlp-1.22}.

\bibitem[Qian et~al.(2024)Qian, Liu, Liu, Chen, Dang, Li, Yang, Chen, Su, Cong, Xu, Li, Liu, and Sun]{qian-etal-2024-chatdev}
Qian, C., Liu, W., Liu, H., Chen, N., Dang, Y., Li, J., Yang, C., Chen, W., Su, Y., Cong, X., Xu, J., Li, D., Liu, Z., and Sun, M.
\newblock {{C}hat{D}ev: Communicative Agents for Software Development}.
\newblock In Ku, L.-W., Martins, A., and Srikumar, V. (eds.), \emph{Proceedings of the 62nd Annual Meeting of the Association for Computational Linguistics (Volume 1: Long Papers)}, pp.\  15174--15186, Bangkok, Thailand, August 2024. Association for Computational Linguistics.
\newblock \doi{10.18653/v1/2024.acl-long.810}.
\newblock URL \url{https://aclanthology.org/2024.acl-long.810/}.

\bibitem[Qiu et~al.(2024)Qiu, Ma, Zhang, and Zhao]{qiu2024megen}
Qiu, J., Ma, X., Zhang, Z., and Zhao, H.
\newblock {MEGen: Generative Backdoor in Large Language Models via Model Editing}.
\newblock \emph{arXiv preprint arXiv:2408.10722}, 2024.

\bibitem[Rafailov et~al.(2024)Rafailov, Sharma, Mitchell, Manning, Ermon, and Finn]{rafailov2024direct}
Rafailov, R., Sharma, A., Mitchell, E., Manning, C.~D., Ermon, S., and Finn, C.
\newblock {Direct Preference Optimization: Your Language Model is Secretly a Reward Model}.
\newblock \emph{Advances in Neural Information Processing Systems}, 36, 2024.

\bibitem[Razdaibiedina et~al.(2023)Razdaibiedina, Mao, Khabsa, Lewis, Hou, Ba, and Almahairi]{razdaibiedina-etal-2023-residual}
Razdaibiedina, A., Mao, Y., Khabsa, M., Lewis, M., Hou, R., Ba, J., and Almahairi, A.
\newblock {Residual Prompt Tuning: Improving Prompt Tuning with Residual Reparameterization}.
\newblock In Rogers, A., Boyd-Graber, J., and Okazaki, N. (eds.), \emph{Findings of the Association for Computational Linguistics: ACL 2023}, pp.\  6740--6757, Toronto, Canada, July 2023. Association for Computational Linguistics.
\newblock \doi{10.18653/v1/2023.findings-acl.421}.
\newblock URL \url{https://aclanthology.org/2023.findings-acl.421/}.

\bibitem[Roberts et~al.(2020)Roberts, Raffel, and Shazeer]{roberts-etal-2020-much}
Roberts, A., Raffel, C., and Shazeer, N.
\newblock {How Much Knowledge Can You Pack Into the Parameters of a Language Model?}
\newblock In Webber, B., Cohn, T., He, Y., and Liu, Y. (eds.), \emph{Proceedings of the 2020 Conference on Empirical Methods in Natural Language Processing (EMNLP)}, pp.\  5418--5426, Online, November 2020. Association for Computational Linguistics.
\newblock \doi{10.18653/v1/2020.emnlp-main.437}.
\newblock URL \url{https://aclanthology.org/2020.emnlp-main.437/}.

\bibitem[Rozner et~al.(2024)Rozner, Battash, Wolf, and Lindenbaum]{rozner2024knowledge}
Rozner, A., Battash, B., Wolf, L., and Lindenbaum, O.
\newblock {Knowledge Editing in Language Models via Adapted Direct Preference Optimization}.
\newblock In Al-Onaizan, Y., Bansal, M., and Chen, Y.-N. (eds.), \emph{Findings of the Association for Computational Linguistics: EMNLP 2024}, pp.\  4761--4774, Miami, Florida, USA, November 2024. Association for Computational Linguistics.
\newblock \doi{10.18653/v1/2024.findings-emnlp.273}.
\newblock URL \url{https://aclanthology.org/2024.findings-emnlp.273/}.

\bibitem[{Secretary of State of California}(2024)]{california_sb942_2024}
{Secretary of State of California}.
\newblock An act to add chapter 25 (commencing with section 22757) to division 8 of the business and professions code, relating to consumer protection, 2024.
\newblock URL \url{https://leginfo.legislature.ca.gov/faces/billNavClient.xhtml?bill_id=202320240SB942}.
\newblock Approved on 19 Sep 2024.

\bibitem[Tan et~al.(2024)Tan, Zhang, and Fu]{tan23malmen}
Tan, C., Zhang, G., and Fu, J.
\newblock {Massive Editing for Large Language Models via Meta Learning}.
\newblock In \emph{International Conference on Learning Representations}, 2024.
\newblock URL \url{https://openreview.net/pdf?id=L6L1CJQ2PE}.

\bibitem[Valipour et~al.(2023)Valipour, Rezagholizadeh, Kobyzev, and Ghodsi]{valipour2022dylora}
Valipour, M., Rezagholizadeh, M., Kobyzev, I., and Ghodsi, A.
\newblock {{D}y{L}o{RA}: Parameter-Efficient Tuning of Pre-trained Models using Dynamic Search-Free Low-Rank Adaptation}.
\newblock In Vlachos, A. and Augenstein, I. (eds.), \emph{Proceedings of the 17th Conference of the European Chapter of the Association for Computational Linguistics}, pp.\  3274--3287, Dubrovnik, Croatia, May 2023. Association for Computational Linguistics.
\newblock \doi{10.18653/v1/2023.eacl-main.239}.
\newblock URL \url{https://aclanthology.org/2023.eacl-main.239/}.

\bibitem[Van~Veen et~al.(2024)Van~Veen, Van~Uden, Blankemeier, Delbrouck, Aali, Bluethgen, Pareek, Polacin, Reis, Seehofnerov{\'a}, et~al.]{van2024adapted}
Van~Veen, D., Van~Uden, C., Blankemeier, L., Delbrouck, J.-B., Aali, A., Bluethgen, C., Pareek, A., Polacin, M., Reis, E.~P., Seehofnerov{\'a}, A., et~al.
\newblock {Adapted Large Language Models Can Outperform Medical Experts in Clinical Text Summarization}.
\newblock \emph{Nature medicine}, 30\penalty0 (4):\penalty0 1134--1142, 2024.

\bibitem[Venditti et~al.(2024)Venditti, Ruzzetti, Xompero, Giannone, Favalli, Romagnoli, and Zanzotto]{venditti2024enhancingdataprivacylarge}
Venditti, D., Ruzzetti, E.~S., Xompero, G.~A., Giannone, C., Favalli, A., Romagnoli, R., and Zanzotto, F.~M.
\newblock {Enhancing Data Privacy in Large Language Models through Private Association Editing}, 2024.
\newblock URL \url{https://arxiv.org/abs/2406.18221}.

\bibitem[Vig et~al.(2020)Vig, Gehrmann, Belinkov, Qian, Nevo, Singer, and Shieber]{vig2020investigating}
Vig, J., Gehrmann, S., Belinkov, Y., Qian, S., Nevo, D., Singer, Y., and Shieber, S.
\newblock {Investigating Gender Bias in Language Models Using Causal Mediation Analysis}.
\newblock \emph{Advances in Neural Information Processing Systems}, 33:\penalty0 12388--12401, 2020.

\bibitem[Wang et~al.(2024{\natexlab{a}})Wang, Lange, Adel, Str{\"o}tgen, and Schuetze]{wang-etal-2024-better}
Wang, M., Lange, L., Adel, H., Str{\"o}tgen, J., and Schuetze, H.
\newblock {Better Call SAUL: Fluent and Consistent Language Model Editing with Generation Regularization}.
\newblock In Al-Onaizan, Y., Bansal, M., and Chen, Y.-N. (eds.), \emph{Findings of the Association for Computational Linguistics: EMNLP 2024}, pp.\  7990--8000, Miami, Florida, USA, November 2024{\natexlab{a}}. Association for Computational Linguistics.
\newblock \doi{10.18653/v1/2024.findings-emnlp.469}.
\newblock URL \url{https://aclanthology.org/2024.findings-emnlp.469/}.

\bibitem[Wang et~al.(2024{\natexlab{b}})Wang, Zhang, Xu, Xi, Deng, Yao, Zhang, Yang, Wang, and Chen]{wang-etal-2024-detoxifying}
Wang, M., Zhang, N., Xu, Z., Xi, Z., Deng, S., Yao, Y., Zhang, Q., Yang, L., Wang, J., and Chen, H.
\newblock {Detoxifying Large Language Models via Knowledge Editing}.
\newblock In Ku, L.-W., Martins, A., and Srikumar, V. (eds.), \emph{Proceedings of the 62nd Annual Meeting of the Association for Computational Linguistics (Volume 1: Long Papers)}, pp.\  3093--3118, Bangkok, Thailand, August 2024{\natexlab{b}}. Association for Computational Linguistics.
\newblock \doi{10.18653/v1/2024.acl-long.171}.
\newblock URL \url{https://aclanthology.org/2024.acl-long.171/}.

\bibitem[Wang et~al.(2024{\natexlab{c}})Wang, Li, Zhang, Xu, Yao, Jiang, Xie, Huang, and Chen]{wang2024wise}
Wang, P., Li, Z., Zhang, N., Xu, Z., Yao, Y., Jiang, Y., Xie, P., Huang, F., and Chen, H.
\newblock {WISE: Rethinking the Knowledge Memory for Lifelong Model Editing of Large Language Models}.
\newblock In \emph{The Thirty-eighth Annual Conference on Neural Information Processing Systems}, 2024{\natexlab{c}}.
\newblock URL \url{https://openreview.net/forum?id=VJMYOfJVC2}.

\bibitem[Wang et~al.(2024{\natexlab{d}})Wang, Zhang, Tian, Xi, Yao, Xu, Wang, Mao, Wang, Cheng, Liu, Ni, Zheng, and Chen]{wang-etal-2024-easyedit}
Wang, P., Zhang, N., Tian, B., Xi, Z., Yao, Y., Xu, Z., Wang, M., Mao, S., Wang, X., Cheng, S., Liu, K., Ni, Y., Zheng, G., and Chen, H.
\newblock {{EasyEdit: An Easy-to-use Knowledge Editing Framework for Large Language Models}}.
\newblock In Cao, Y., Feng, Y., and Xiong, D. (eds.), \emph{Proceedings of the 62nd Annual Meeting of the Association for Computational Linguistics (Volume 3: System Demonstrations)}, pp.\  82--93, Bangkok, Thailand, August 2024{\natexlab{d}}. Association for Computational Linguistics.
\newblock \doi{10.18653/v1/2024.acl-demos.9}.
\newblock URL \url{https://aclanthology.org/2024.acl-demos.9/}.

\bibitem[Wang et~al.(2024{\natexlab{e}})Wang, Mao, Wu, Ge, Wei, and Ji]{wang-etal-2024-unleashing}
Wang, Z., Mao, S., Wu, W., Ge, T., Wei, F., and Ji, H.
\newblock {Unleashing the Emergent Cognitive Synergy in Large Language Models: A Task-Solving Agent through Multi-Persona Self-Collaboration}.
\newblock In Duh, K., Gomez, H., and Bethard, S. (eds.), \emph{Proceedings of the 2024 Conference of the North American Chapter of the Association for Computational Linguistics: Human Language Technologies (Volume 1: Long Papers)}, pp.\  257--279, Mexico City, Mexico, June 2024{\natexlab{e}}. Association for Computational Linguistics.
\newblock \doi{10.18653/v1/2024.naacl-long.15}.
\newblock URL \url{https://aclanthology.org/2024.naacl-long.15/}.

\bibitem[Wester et~al.(2024)Wester, De~Jong, Pohl, and Van~Berkel]{wester2024exploring}
Wester, J., De~Jong, S., Pohl, H., and Van~Berkel, N.
\newblock {Exploring People’s Perceptions of LLM-generated Advice}.
\newblock \emph{Computers in Human Behavior: Artificial Humans}, pp.\  100072, 2024.

\bibitem[Wortsman et~al.(2022)Wortsman, Ilharco, Gadre, Roelofs, Gontijo-Lopes, Morcos, Namkoong, Farhadi, Carmon, Kornblith, et~al.]{wortsman2022model}
Wortsman, M., Ilharco, G., Gadre, S.~Y., Roelofs, R., Gontijo-Lopes, R., Morcos, A.~S., Namkoong, H., Farhadi, A., Carmon, Y., Kornblith, S., et~al.
\newblock {Model Soups: Averaging Weights of Multiple Fine-tuned Models Improves Accuracy without Increasing Inference Time}.
\newblock In \emph{International Conference on Machine Learning}, pp.\  23965--23998. PMLR, 2022.

\bibitem[Xi et~al.(2023)Xi, Chen, Guo, He, Ding, Hong, Zhang, Wang, Jin, Zhou, Zheng, Fan, Wang, Xiong, Zhou, Wang, Jiang, Zou, Liu, Yin, Dou, Weng, Cheng, Zhang, Qin, Zheng, Qiu, Huang, and Gui]{xi2023rise}
Xi, Z., Chen, W., Guo, X., He, W., Ding, Y., Hong, B., Zhang, M., Wang, J., Jin, S., Zhou, E., Zheng, R., Fan, X., Wang, X., Xiong, L., Zhou, Y., Wang, W., Jiang, C., Zou, Y., Liu, X., Yin, Z., Dou, S., Weng, R., Cheng, W., Zhang, Q., Qin, W., Zheng, Y., Qiu, X., Huang, X., and Gui, T.
\newblock {The Rise and Potential of Large Language Model Based Agents: A Survey}, 2023.
\newblock arXiv:2309.07864.

\bibitem[Yang et~al.(2024)Yang, Sun, Tan, Ma, Su, Yin, and Shen]{yang-etal-2024-fall}
Yang, W., Sun, F., Tan, J., Ma, X., Su, D., Yin, D., and Shen, H.
\newblock {The Fall of ROME: Understanding the Collapse of LLMs in Model Editing}.
\newblock In Al-Onaizan, Y., Bansal, M., and Chen, Y.-N. (eds.), \emph{Findings of the Association for Computational Linguistics: EMNLP 2024}, pp.\  4079--4087, Miami, Florida, USA, November 2024. Association for Computational Linguistics.
\newblock \doi{10.18653/v1/2024.findings-emnlp.236}.
\newblock URL \url{https://aclanthology.org/2024.findings-emnlp.236/}.

\bibitem[Youssef et~al.(2023)Youssef, Kora{\c{s}}, Li, Schl{\"o}tterer, and Seifert]{youssef-etal-2023-give}
Youssef, P., Kora{\c{s}}, O., Li, M., Schl{\"o}tterer, J., and Seifert, C.
\newblock {Give Me the Facts! A Survey on Factual Knowledge Probing in Pre-trained Language Models}.
\newblock In Bouamor, H., Pino, J., and Bali, K. (eds.), \emph{Findings of the Association for Computational Linguistics: EMNLP 2023}, pp.\  15588--15605, Singapore, December 2023. Association for Computational Linguistics.
\newblock \doi{10.18653/v1/2023.findings-emnlp.1043}.
\newblock URL \url{https://aclanthology.org/2023.findings-emnlp.1043/}.

\bibitem[Youssef et~al.(2025{\natexlab{a}})Youssef, Zhao, Schl{\"o}tterer, and Seifert]{youssef-etal-2025-make}
Youssef, P., Zhao, Z., Schl{\"o}tterer, J., and Seifert, C.
\newblock {How to Make LLMs Forget: On Reversing In-Context Knowledge Edits}.
\newblock In Chiruzzo, L., Ritter, A., and Wang, L. (eds.), \emph{Proceedings of the 2025 Conference of the Nations of the Americas Chapter of the Association for Computational Linguistics: Human Language Technologies (Volume 1: Long Papers)}, pp.\  12656--12669, Albuquerque, New Mexico, April 2025{\natexlab{a}}. Association for Computational Linguistics.
\newblock ISBN 979-8-89176-189-6.
\newblock URL \url{https://aclanthology.org/2025.naacl-long.630/}.

\bibitem[Youssef et~al.(2025{\natexlab{b}})Youssef, Zhao, Seifert, and Schl{\"o}tterer]{youssef-etal-2025-fact}
Youssef, P., Zhao, Z., Seifert, C., and Schl{\"o}tterer, J.
\newblock {Has this Fact been Edited? Detecting Knowledge Edits in Language Models}.
\newblock In Chiruzzo, L., Ritter, A., and Wang, L. (eds.), \emph{Proceedings of the 2025 Conference of the Nations of the Americas Chapter of the Association for Computational Linguistics: Human Language Technologies (Volume 1: Long Papers)}, pp.\  9768--9784, Albuquerque, New Mexico, April 2025{\natexlab{b}}. Association for Computational Linguistics.
\newblock ISBN 979-8-89176-189-6.
\newblock URL \url{https://aclanthology.org/2025.naacl-long.492/}.

\bibitem[Yu et~al.(2024)Yu, Chen, Zhou, and He]{yu2023melo}
Yu, L., Chen, Q., Zhou, J., and He, L.
\newblock {MELO: Enhancing Model Editing with Neuron-Indexed Dynamic LoRA }.
\newblock AAAI'24/IAAI'24/EAAI'24. AAAI Press, 2024.
\newblock ISBN 978-1-57735-887-9.
\newblock \doi{10.1609/aaai.v38i17.29916}.
\newblock URL \url{https://doi.org/10.1609/aaai.v38i17.29916}.

\bibitem[Zhang et~al.(2024{\natexlab{a}})Zhang, Tian, Cheng, Liang, Hu, Xue, Gou, Chen, and Chen]{ijcai2024p0733}
Zhang, N., Tian, B., Cheng, S., Liang, X., Hu, Y., Xue, K., Gou, Y., Chen, X., and Chen, H.
\newblock {InstructEdit: Instruction-Based Knowledge Editing for Large Language Models}.
\newblock In Larson, K. (ed.), \emph{Proceedings of the Thirty-Third International Joint Conference on Artificial Intelligence, {IJCAI-24}}, pp.\  6633--6641. International Joint Conferences on Artificial Intelligence Organization, 8 2024{\natexlab{a}}.
\newblock \doi{10.24963/ijcai.2024/733}.
\newblock URL \url{https://doi.org/10.24963/ijcai.2024/733}.
\newblock Main Track.

\bibitem[Zhang et~al.(2024{\natexlab{b}})Zhang, Yao, Tian, Wang, Deng, Wang, Xi, Mao, Zhang, Ni, et~al.]{zhang2024comprehensive}
Zhang, N., Yao, Y., Tian, B., Wang, P., Deng, S., Wang, M., Xi, Z., Mao, S., Zhang, J., Ni, Y., et~al.
\newblock {A Comprehensive Study of Knowledge Editing for Large Language Models}.
\newblock \emph{arXiv preprint arXiv:2401.01286}, 2024{\natexlab{b}}.

\bibitem[Zhang et~al.(2023)Zhang, Tan, Xu, Wang, Huang, and Huang]{zhang2023towards}
Zhang, Z.-R., Tan, C., Xu, H., Wang, C., Huang, J., and Huang, S.
\newblock {Towards Adaptive Prefix Tuning for Parameter-Efficient Language Model Fine-tuning}.
\newblock In Rogers, A., Boyd-Graber, J., and Okazaki, N. (eds.), \emph{Proceedings of the 61st Annual Meeting of the Association for Computational Linguistics (Volume 2: Short Papers)}, pp.\  1239--1248, Toronto, Canada, July 2023. Association for Computational Linguistics.
\newblock \doi{10.18653/v1/2023.acl-short.107}.
\newblock URL \url{https://aclanthology.org/2023.acl-short.107/}.

\bibitem[Zhao et~al.(2024)Zhao, Li, Li, Zhang, and Sun]{zhao-etal-2024-defending-large}
Zhao, W., Li, Z., Li, Y., Zhang, Y., and Sun, J.
\newblock {Defending Large Language Models Against Jailbreak Attacks via Layer-specific Editing}.
\newblock In Al-Onaizan, Y., Bansal, M., and Chen, Y.-N. (eds.), \emph{Findings of the Association for Computational Linguistics: EMNLP 2024}, pp.\  5094--5109, Miami, Florida, USA, November 2024. Association for Computational Linguistics.
\newblock \doi{10.18653/v1/2024.findings-emnlp.293}.
\newblock URL \url{https://aclanthology.org/2024.findings-emnlp.293/}.

\bibitem[Zheng et~al.(2023)Zheng, Li, Dong, Fan, Wu, Xu, and Chang]{zheng-etal-2023-edit}
Zheng, C., Li, L., Dong, Q., Fan, Y., Wu, Z., Xu, J., and Chang, B.
\newblock {Can We Edit Factual Knowledge by In-Context Learning?}
\newblock In Bouamor, H., Pino, J., and Bali, K. (eds.), \emph{Proceedings of the 2023 Conference on Empirical Methods in Natural Language Processing}, pp.\  4862--4876, Singapore, December 2023. Association for Computational Linguistics.
\newblock \doi{10.18653/v1/2023.emnlp-main.296}.
\newblock URL \url{https://aclanthology.org/2023.emnlp-main.296/}.

\end{thebibliography}
\bibliographystyle{icml2025}

\newpage

\appendix

\onecolumn

\end{document}